\documentclass[lettersize,journal]{IEEEtran}

\usepackage{amsmath,amsfonts}
\usepackage{algorithmic}
\usepackage{algorithm}
\usepackage{array}
\usepackage{textcomp}
\usepackage{stfloats}
\usepackage{url}
\usepackage{verbatim}
\usepackage{graphicx}
\usepackage{cite}
\usepackage{graphicx}
\usepackage{amsmath}
\usepackage{graphics}
\usepackage{amsfonts}
\usepackage{amssymb}%
\usepackage{amsthm}
\usepackage{hyperref}
\usepackage{epsfig}
\usepackage{subfigure}
\usepackage{color}
\usepackage{listings}
\usepackage{booktabs}
\usepackage{multirow}
\usepackage{algorithm}

\usepackage{algorithmic}
\usepackage{amsfonts}
\usepackage{multicol}
\usepackage{epstopdf}

\usepackage{cleveref}

\begin{document}




\title{Reducing Capacity Gap in Knowledge Distillation with Review Mechanism for Crowd Counting}



\author{Yunxin Liu, Qiaosi Yi, and Jinshan Zeng
\IEEEcompsocitemizethanks{\IEEEcompsocthanksitem Y. Liu and J. Zeng are with the School of Computer and Information Engineering, Jiangxi Normal University, Nanchang, China (email: liudearbreeze@gmail.com (Y. Liu), jinshanzeng@jxnu.edu.cn (J. Zeng)).
\IEEEcompsocthanksitem Q. Yi is with the School of Information Science and Engineering, East China University of Science and Technology, Shanghai, China (email: qiaosiyijoyies@gmail.com).
}
\thanks{The corresponding author is Jinshan Zeng (jinshanzeng@jxnu.edu.cn)}
}

\maketitle

\begin{abstract}
The lightweight crowd counting models, in particular knowledge distillation (KD) based models, have attracted rising attention in recent years due to their superiority on computational efficiency and hardware requirement.
However, existing KD based models usually suffer from the capacity gap issue, resulting in the performance of the student network being limited by the teacher network. 
In this paper, we address this issue by introducing a novel review mechanism following KD models, motivated by the review mechanism of human-beings during the study.
Thus, the proposed model is dubbed \textit{ReviewKD}. The proposed model consists of an instruction phase and a review phase, where we firstly exploit a well-trained heavy teacher network to transfer its latent feature to a lightweight student network in the instruction phase,
then in the review phase yield a refined estimate of the density map based on the learned feature through a review mechanism.
In the review mechanism, we firstly use the estimate of the density map yielded by the decoder as a certain attention weight to pay more attention to the important crowd regions, then aggregates the new feature incorporated with attention weight and the feature learned from the teacher to generate an enhanced feature, and finally yield a refined estimate by the decoder. 
The effectiveness of ReviewKD is demonstrated by a set of experiments over six benchmark datasets via comparing to the state-of-the-art models. Numerical results show that ReviewKD outperforms existing lightweight models for crowd counting, and can effectively alleviate the capacity gap issue, and particularly has the performance beyond the teacher network. Besides the lightweight models, we also show that the suggested review mechanism can be used as a plug-and-play module to further boost the performance of a kind of heavy crowd counting models without modifying the neural network architecture and introducing any additional model parameter.
\end{abstract}


\section{Introduction}

\IEEEPARstart{C}{rowd} counting aims to automatically estimate the total number of people in surveillance scenes. It is of great significance in many fields such as security analysis, and has attracted tremendous attention in recent years \cite{ma2019bayesian,jiang2020attention,thanasutives2021encoder}.
	
Existing methods for crowd counting can be generally divided into three categories, i.e.,  detector based methods \cite{topkaya2014counting,li2008estimating,leibe2005pedestrian,enzweiler2008monocular}, regressor based methods \cite{chan2008privacy,idrees2013multi,chan2009bayesian,chen2012feature} and  deep learning based methods \cite{bai2020adaptive,zhang2016single,li2018csrnet,miao2020shallow,sindagi2019ha,walach2016learning,wang2015deep,fu2015fast,zhang2015cross,yan2021crowd}.
The first category of crowd counting methods generally yields the estimate of crowd count by locating the people with pedestrian detectors, which mainly focus on some important characteristics of human-beings such as face and body \cite{leibe2005pedestrian,dollar2011pedestrian}. These methods are usually effective when applied to sparse scenes with a small number of objectives, but may fail for those scenes with complex background and dense objectives, and in particular with crowd occlusion. Later, some regressor based methods  were suggested in the literature \cite{chan2008privacy,idrees2013multi,chan2009bayesian,chen2012feature}. This kind of methods yields the estimate of crowd count via learning a mapping from some handcrafted low-level features. Although  the regressor based methods can in some extent alleviate the problems of occlusion and background clutter, yet their performance depends on the quality of low-level features and is usually far from satisfactory for real-world applications.

Motivated by the powerful expression and approximation ability of deep neural networks, in particular the great success of convolutional neural networks (CNNs), the deep learning based methods have become the mainstream in crowd counting and made remarkable progress \cite{bai2020adaptive,zhang2016single,chen2020crowd,li2018csrnet,jiang2020density,miao2020shallow,wang2021interlayer,sindagi2019ha,walach2016learning,wang2015deep,fu2015fast,zhang2015cross}.
To achieve better performance, most of state-of-the-art methods \cite{bai2020adaptive,zhang2016single,li2018csrnet,jiang2020density,miao2020shallow,sindagi2019ha,walach2016learning,wang2015deep,fu2015fast,zhang2015cross} use heavy backbone networks such as the VGG networks \cite{simonyan2014very} as feature extractors.
Although these heavy crowd counting models can achieve satisfactory performance on estimating the crowd count, their impressive performance is	at the expensive of large computational cost and hardware burden \cite{liu2020efficient}, limiting their wide use in the real-world applications and causing poor scalability, particularly, on edge computing devices\footnote{\url{https://en.wikipedia.org/wiki/Edge_computi}} with limited computing resources.

\begin{figure*}[!t]
\begin{minipage}[b]{0.192\linewidth}
\centering
\includegraphics*[scale=.114]{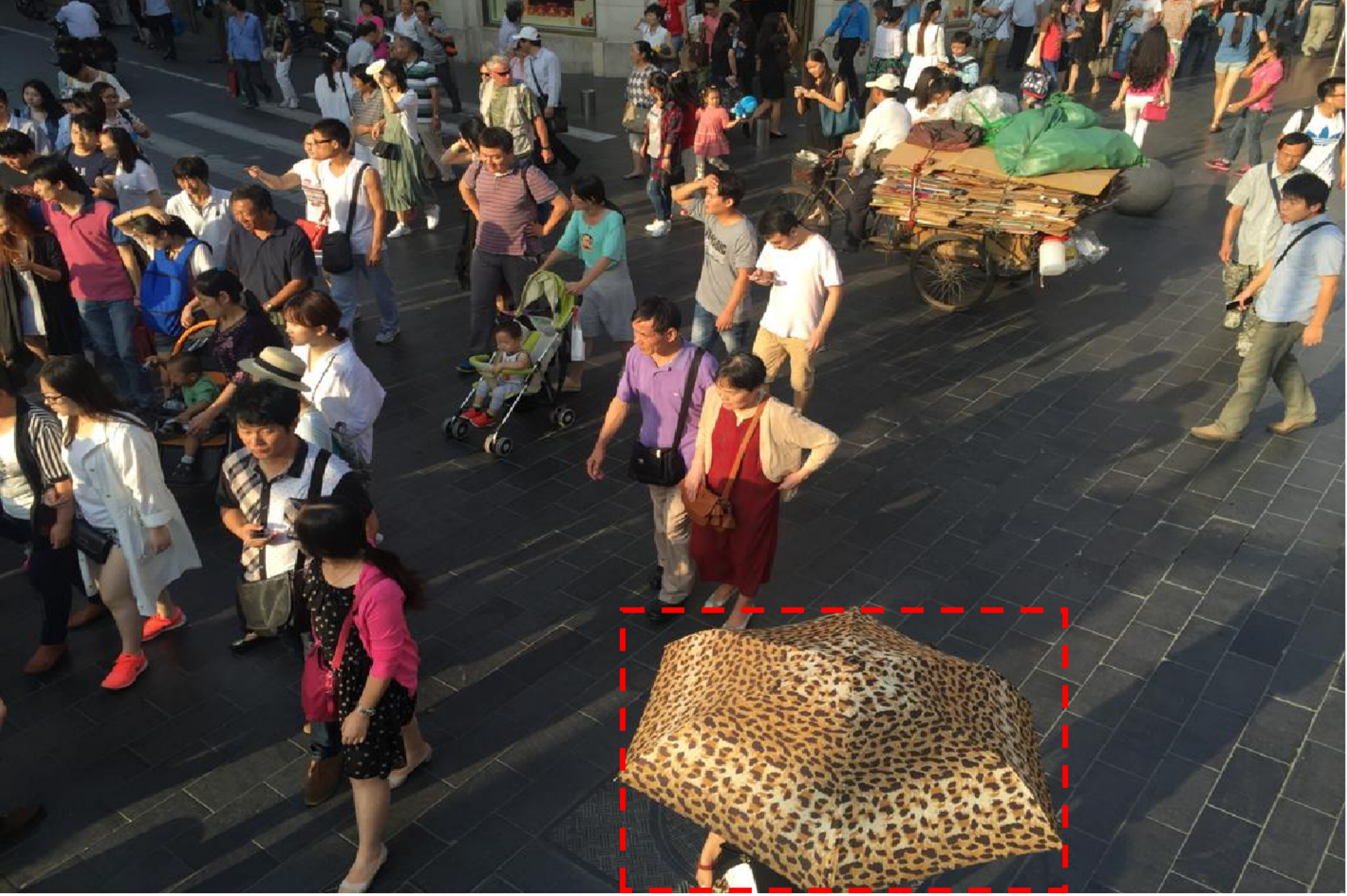}
\centerline{{\footnotesize  (a) Original scene}}
\end{minipage}
\hfill
\begin{minipage}[b]{0.192\linewidth}
\centering
\includegraphics*[scale=.115]{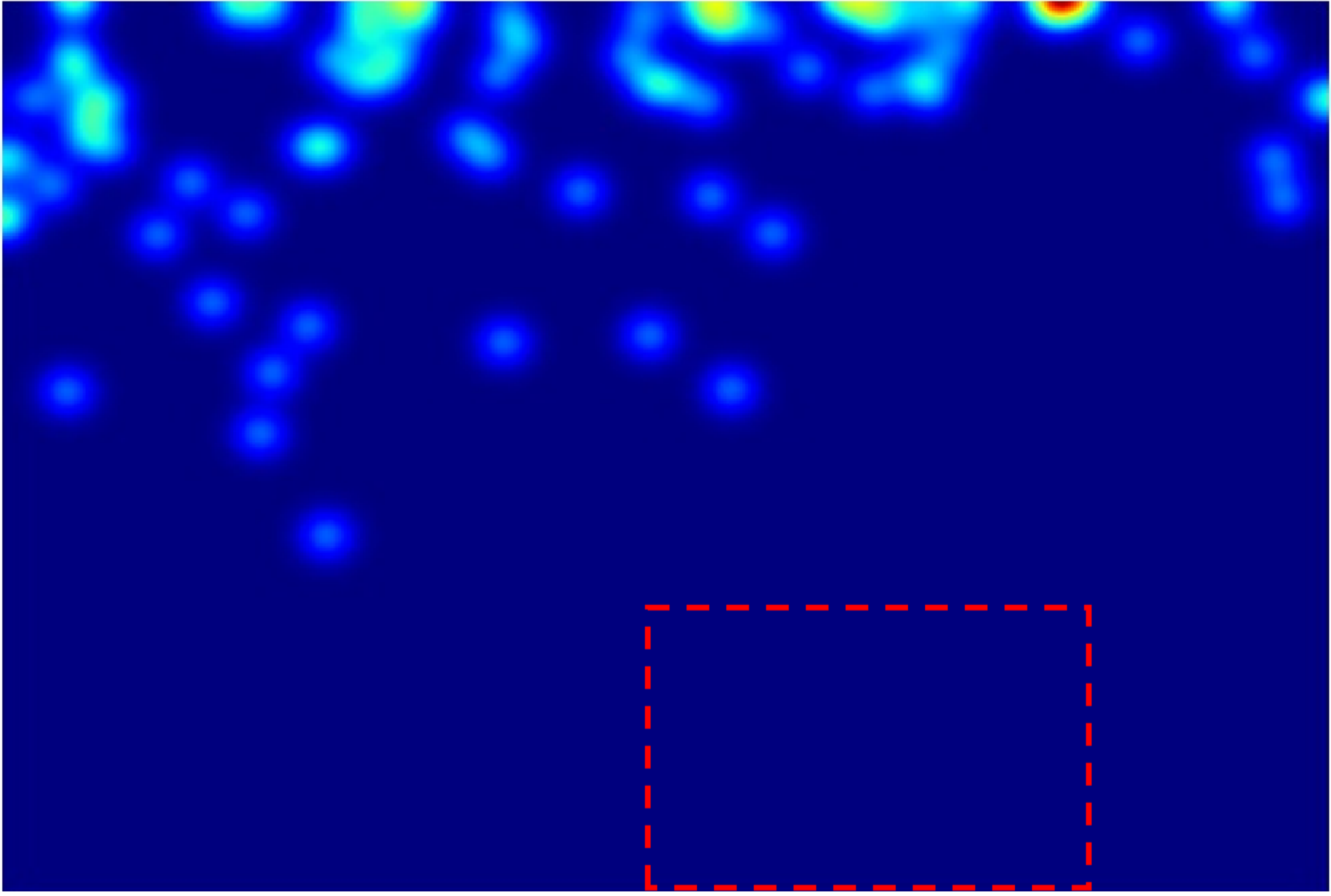}
\centerline{{\footnotesize  (b) Ground truth (count:71)}}
\end{minipage}
\hfill
\begin{minipage}[b]{0.192\linewidth}
\centering
\includegraphics*[scale=.115]{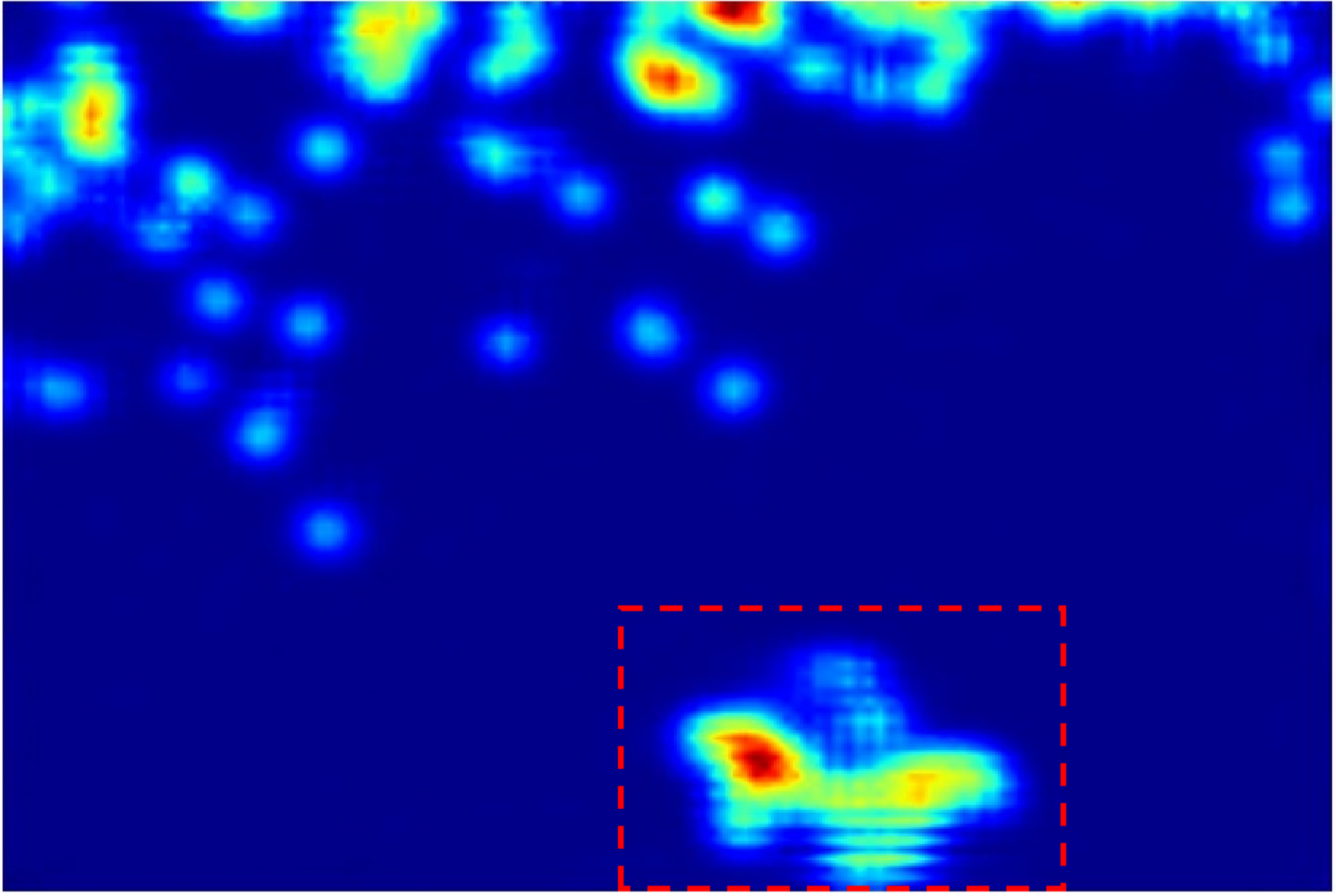}
\centerline{{\footnotesize  (c) Teacher (count: 97)}}
\end{minipage}
\hfill
\begin{minipage}[b]{0.192\linewidth}
\centering
\includegraphics*[scale=.115]{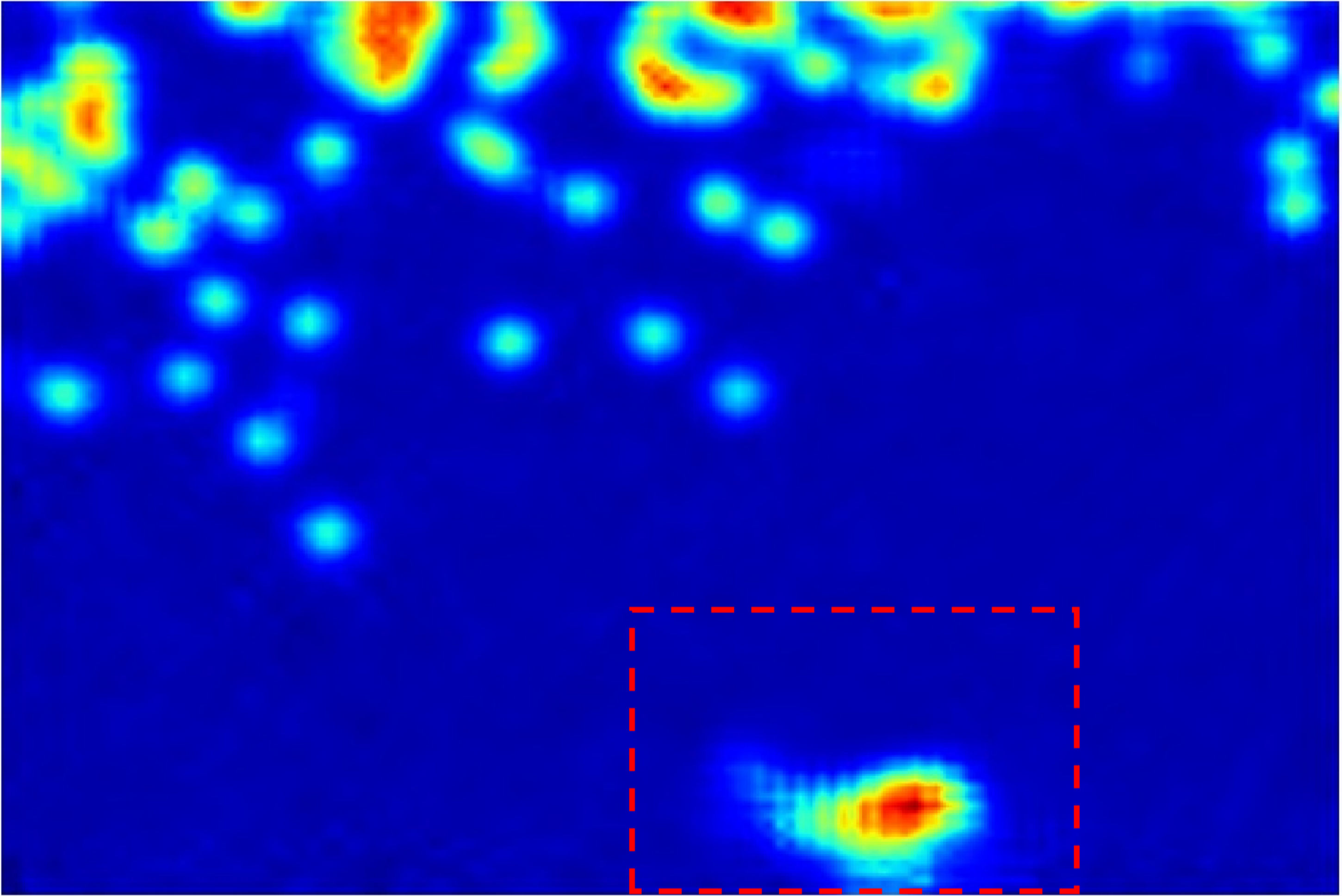}
\centerline{{\footnotesize  (d) SKT (count: 84)}}
\end{minipage}
\hfill
\begin{minipage}[b]{0.192\linewidth}
\centering
\includegraphics*[scale=.115]{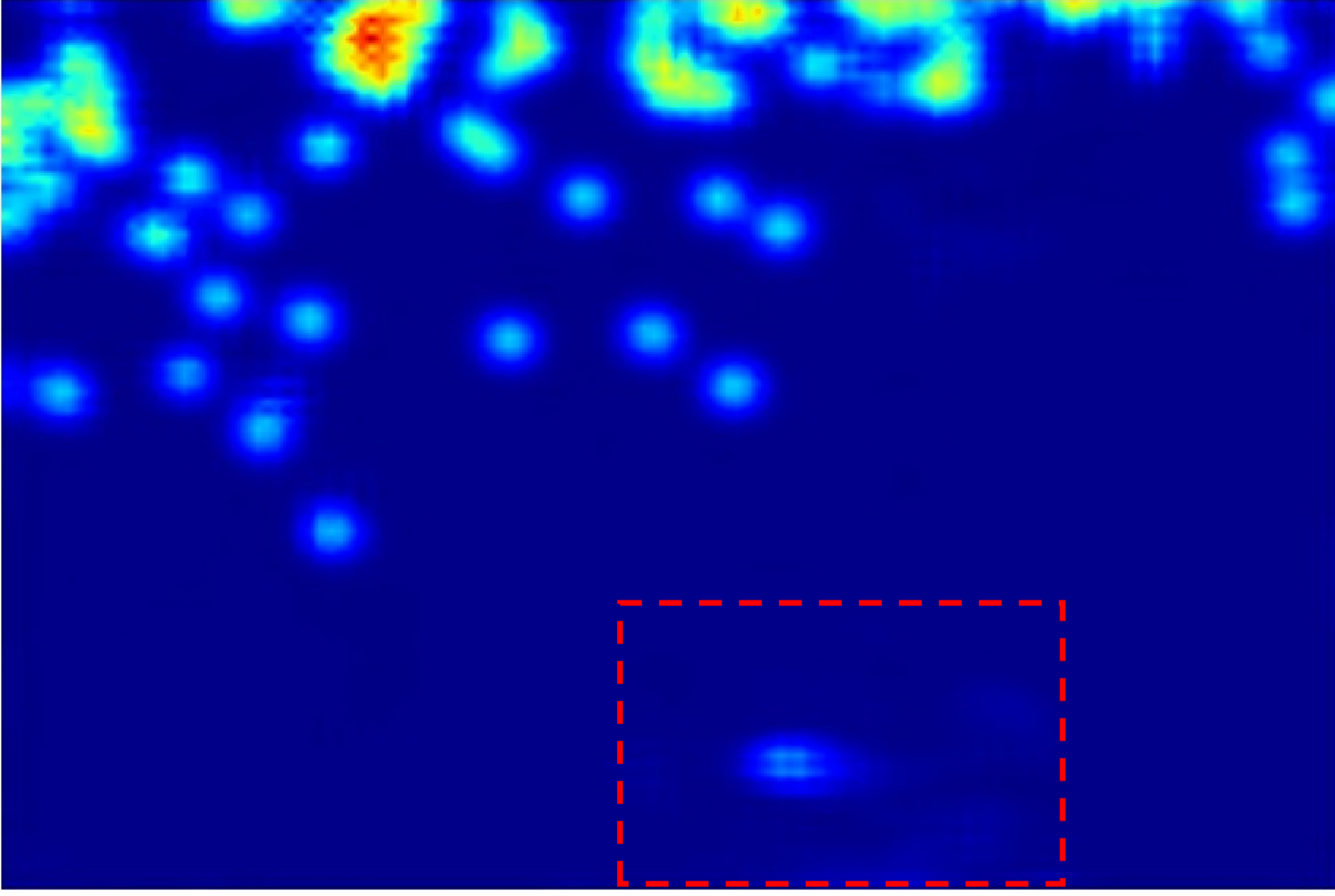}
\centerline{{\footnotesize  (e) VSKT (count:76)}}
\end{minipage}
\hfill
\caption{Alleviating error propagation issue by the suggested VSKT model. (a) The original scene with obscuration by an umbrella; (b)-(e) present the density maps of ground truth, the teacher network, SKT and VSKT, respectively.
}
\label{fig:mistake-propagation}
\end{figure*}

\begin{figure*}[!t]
\begin{minipage}[b]{1\linewidth}
\centering
\includegraphics*[scale=.55]{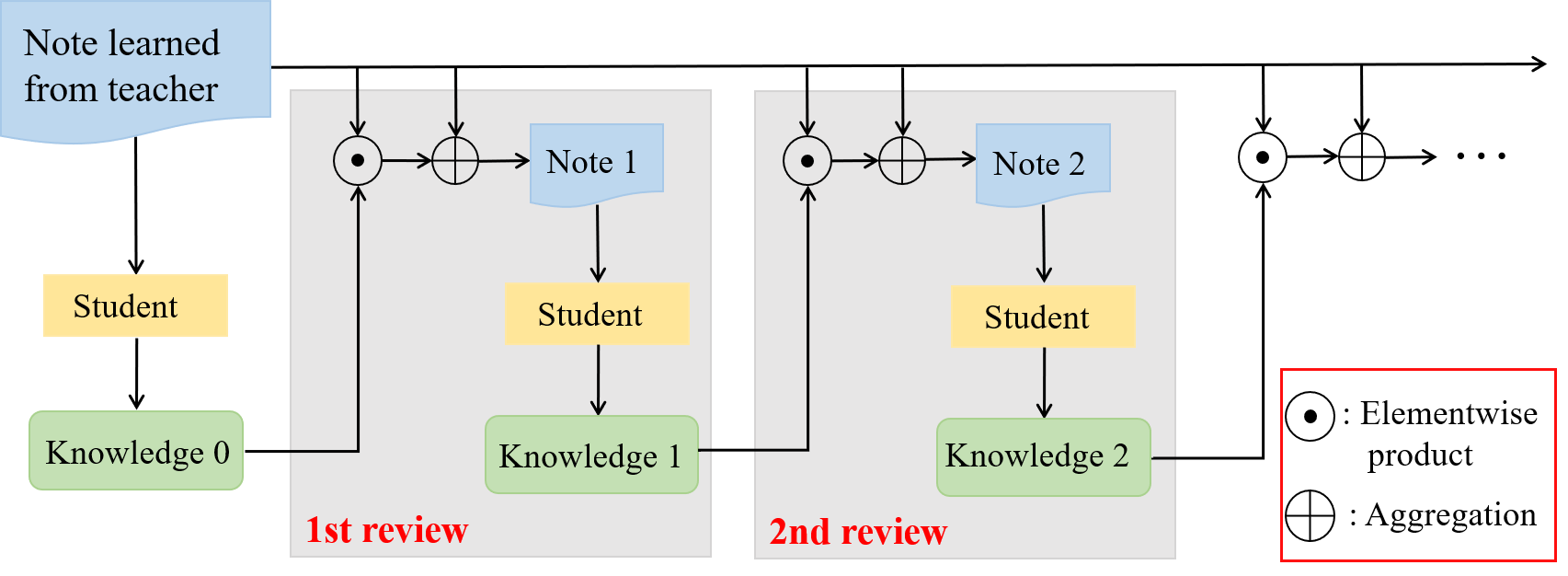}
\end{minipage}
\hfill
\caption{A simplified review mechanism of human-beings. The student refines his knowledge through several rounds of review. 
}
\label{fig:review-mechanism}
\end{figure*}

To fill this gap, several lightweight crowd counting models were proposed in recent years \cite{zhang2016single,cao2018scale,shen2018crowd,jiang2019crowd,wang2020mobilecount}. Although the computational cost is significantly reduced,  the performance of these lightweight models also degrades much, as compared to those heavy models in \cite{bai2020adaptive,zhang2016single,li2018csrnet,miao2020shallow,sindagi2019ha,walach2016learning,wang2015deep,fu2015fast,zhang2015cross}.
Along this line, an efficient lightweight crowd counting model called the structured knowledge transfer (SKT) was recently suggested in the literature \cite{liu2020efficient}, based on the framework of knowledge distillation (KD) \cite{hinton2015distilling}. In the framework of KD, a lightweight student network is trained to acquire the knowledge of a well-trained heavy teacher network. To fully distill the knowledge of the teacher network, two complementary modules, i.e., an Intra-Layer Pattern Transfer (Intra-PT) and an Inter-Layer Relation Pattern (Inter-RT) were introduced in \cite{liu2020efficient} to exploit the structured knowledge of the teacher network. 

Despite the effectiveness of SKT \cite{liu2020efficient}, it may still suffer from the \textit{capacity gap issue}, an essential issue existing in the framework of KD \cite{hinton2015distilling,li2021reskd,gao2020residual,mirzadeh2020improved}, that is, there is a gap between the capacity of the teacher and student networks. This issue generally results in the performance of the student network being limited by the teacher network. Thus, a natural question arisen here is that 	\textit{``how to overcome the capacity gap issue such that the performance of student network can be beyond the associated teacher network?''}	Moreover, we conducted a more in depth analysis of SKT and observed that SKT may also suffer from the \textit{error propagation issue}, that is, errors made by the teacher network may propagate to the student network, as shown in Figure \ref{fig:mistake-propagation}. From Figure \ref{fig:mistake-propagation}(c), the teacher network makes some errors in the obscured region of an umbrella, and then propagates them to the student network, as shown in Figure \ref{fig:mistake-propagation}(d).

To address these issues, we introduce a review mechanism following KD models, mainly motivated by the review mechanism of human-beings during the study as shown in Figure \ref{fig:review-mechanism}. Thus, the suggested model is called \textit{ReviewKD}, whose architecture is presented in Figure \ref{fig:model-architecuture}. There are two phases in the proposed model, i.e, an instruction phase and a review phase. In the instruction phase, a well-trained teacher network transfers its latent feature to a lightweight student network, and in the review phase, a refined estimate of the density map is yielded by the decoder of the student network from the feature aggregated through a review mechanism.
In the review mechanism, we firstly use the density map yielded by the decoder of the student network  as a certain attention map to emphasize the important crowd regions of the feature learned from the teacher network, then aggregate the learned feature and its enhanced one incorporated with attention map to generate a refined feature, and finally  yield a refined density map by the decoder from the aggregated feature. 
By the use of such review mechanism, the capacity gap issue suffered by KD models can be effectively alleviated.
Our major contributions can be summarized as follows.
\begin{enumerate}
	  \item[(1)] We introduce a novel review mechanism to solve the capacity gap issue suffered by existing KD models, mainly motivated by the review mechanism of human-beings. By leveraging such review mechanism, the capacity gap issue can be effectively alleviated, and in particular, the performance of the student network can be generally better than the associated teacher network.
	    
	  \item[(2)] In the instruction phase of the suggested ReviewKD, we consider two KD models including the SKT recently proposed in \cite{liu2020efficient} and its variant suggested in this paper by using a new lightweight student network constructed by MobileNetV2. 
	  The effectiveness of the proposed models is demonstrated by a series of experiments over six benchmark datasets in comparison to the state-of-the-art models for crowd counting. Numerical results show that the proposed models outperform the state-of-the-art models for crowd counting, and the capacity gap issue suffered by KD models can be effectively addressed by the introduced review mechanism. Moreover, the error propagation issue suffered by SKT can be also effectively alleviated by the suggested VSKT as shown in Figure \ref{fig:mistake-propagation}(e). 

	  \item[(3)] We also conduct a set of numerical experiments to show that the introduced review mechanism can be used as a plug-and-play module to further improve the performance of a kind of heavy neural network models (say, the encoder-decoder based models) for crowd counting without modifying the neural network architectures and introducing any additional model parameter. 
\end{enumerate}

The rest of this paper is organized as follows. Section \ref{sc:related-work} presents some related work of this paper. Section \ref{sc:proposed-model} describes the proposed model in detail. Section \ref{sc:experiment} provides a series of experiments to demonstrate the effectiveness of the proposed model. We conclude this paper in Section \ref{sc:conclusion}.

\section{Related Work}	
\label{sc:related-work}

The capacity gap issue existing in knowledge distillation will limit the performance of the kind of KD models, i.e., generally yielding a lightweight student network with worse performance than its associated teacher network \cite{li2021reskd,gao2020residual,mirzadeh2020improved}.
In order to address this issue, some related works have been recently suggested in the literature \cite{mirzadeh2020improved,gao2020residual,li2021reskd} for image classification.
In \cite{mirzadeh2020improved}, a \textit{teacher assistant} was introduced to reduce the capacity gap between the student and teacher networks.
In \cite{gao2020residual}, an \textit{assistant} was introduced to learn the residual error between the knowledge of the student and teacher such that the student gradually learns the knowledge of the teacher.
Similar idea was also suggested in \cite{li2021reskd}, where a \textit{res-student} was introduced to guide the student to learn the knowledge of the teacher.
On one hand, it can be observed that the eixsting literature \cite{mirzadeh2020improved,gao2020residual,li2021reskd} focus on ``\textit{how to yield a student network with performance close to the teacher network}'', while this paper focuses on ``\textit{how to yield a student network with performance beyond the teacher network}''.
To achieve this, we introduce a novel review mechanism to further boost the performance of the student network such that it can be beyond the performance of the teacher network, mainly motivated the review mechanism of human-beings.
Moreover, it should be pointed out that all the assisted modules, i.e., \textit{teacher assistant}, \textit{assistant} and \textit{res-student} suggested in the literature \cite{mirzadeh2020improved,gao2020residual,li2021reskd} need to interact with both student and teacher networks, while the \textit{review} mechanism introduced in this paper only needs to interact with the student network. This makes the introduced review mechanism can be easily adapted to many KD models.
On the other hand, existing work in the literature \cite{mirzadeh2020improved,gao2020residual,li2021reskd} are mainly applicable to image classification, yet it is difficult to directly apply these improved KD models to the more challenging dense-labeling crowd counting, since this task requires the distilled models to maintain the discriminative abilities at every location \cite{liu2020efficient}.

\section{Proposed Model}
\label{sc:proposed-model}
In this section, we firstly describe a simplified review mechanism of human-beings, then motivated by it, introduce the proposed model, and finally present some specific implementations of the proposed model.

\begin{figure*}[!t]
\begin{minipage}[b]{1\linewidth}
\centering
\includegraphics*[scale=.55]{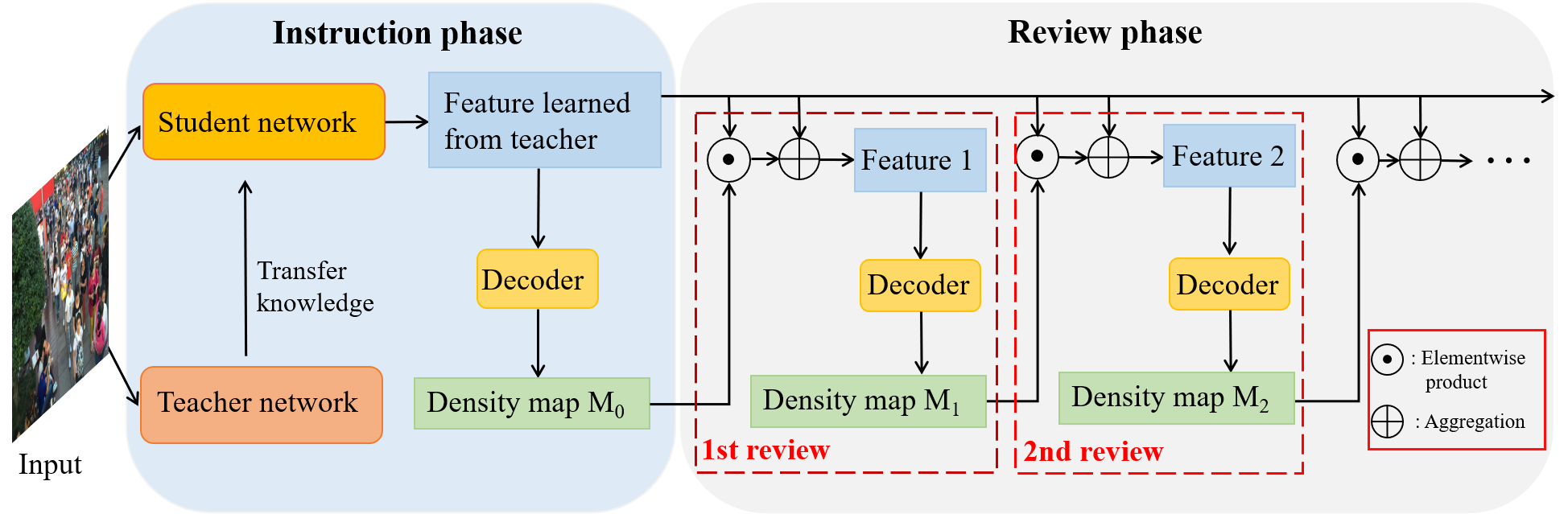}
\end{minipage}
\hfill
\caption{The architecture of the proposed ReviewKD, where a KD model is used in the instruction phase to transfer the knowledge of the heavy teacher network to the lightweight student network, and a review mechanism is introduced to improve the performance of the student network in the review phase, motivated by the review mechanism of human-beings.
}
\label{fig:model-architecuture}
\end{figure*}

\subsection{Review mechanism of human-beings}
As an old Chinese idiom said, ``new understanding can be yielded through reviewing the old knowledge'', thus the review mechanism is a very important self-enhance scheme during the study. A simplified review mechanism can be described in Figure \ref{fig:review-mechanism}. From Figure \ref{fig:review-mechanism}, given the note learned from the teacher and the associated knowledge yielded by the student during the instruction phase, in the first round of review, the student uses the learned knowledge as a certain attention map to pay more attention to the important places in the note, and later aggregates such concerned note after the attention procedure and the original note learned from the teacher to generate a refined note, and finally yields some refined knowledge based on the new note. Such review procedure can be repeated several rounds to produce more new knowledge as shown in Figure \ref{fig:review-mechanism}.

\begin{figure*}[!t]
\begin{minipage}[b]{1\linewidth}
\centering
\includegraphics*[scale=.55]{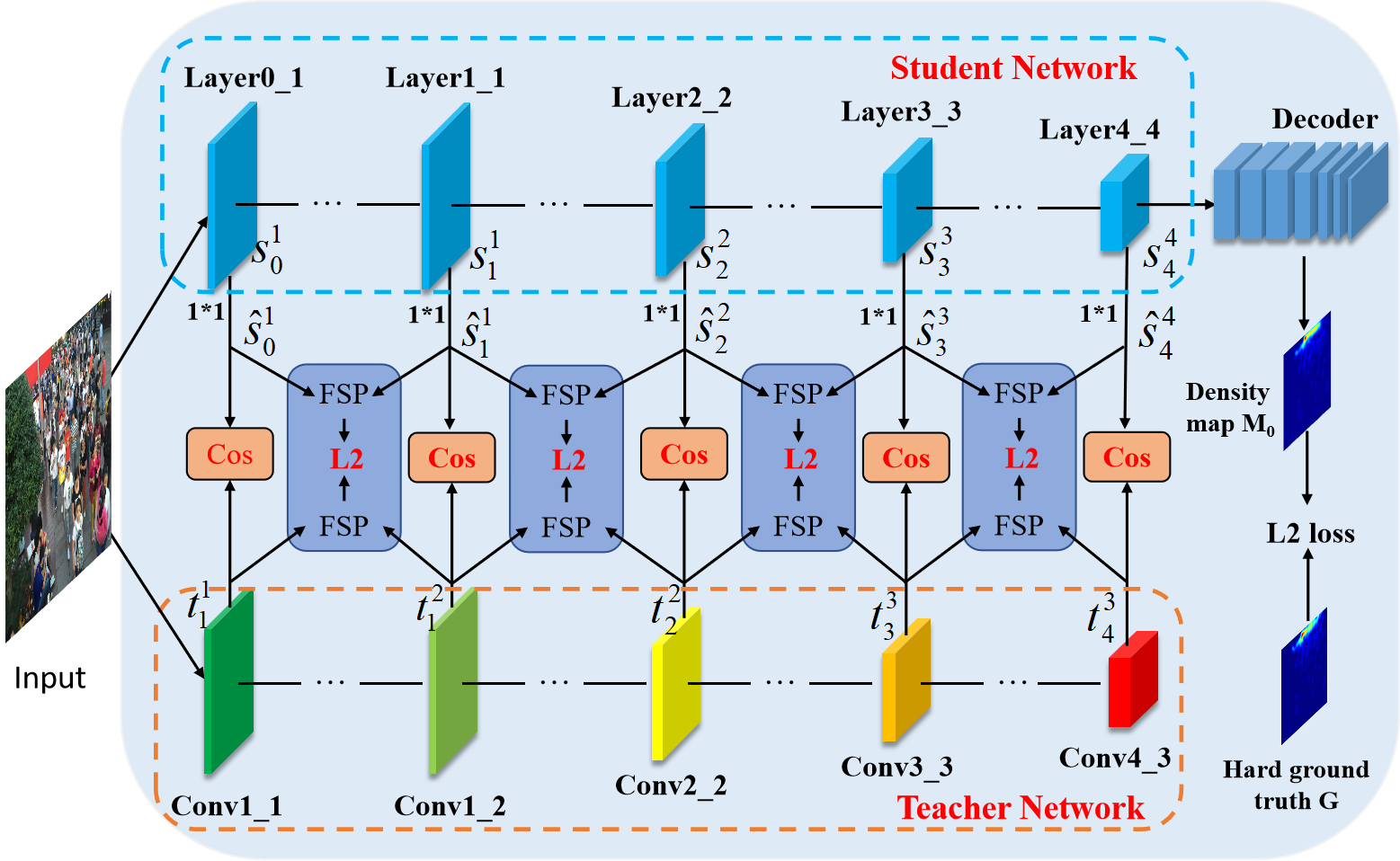}
\end{minipage}
\hfill
\caption{A variant of SKT model introduced in this paper, where the CSRNet \cite{li2018csrnet} is used as the teacher network as an example.
}
\label{fig:vskt-model}
\end{figure*}

\begin{table*}
		\centering
		\caption{The structure of encoder in the student network of VKST. All layers are residual modules from MobileNetV2 \cite{howard2018inverted}. From the second to the fourth column, the values in the table represent expansion factor, number of blocks, and stride, respectively. We consider three different student networks with different $n$, i.e., 1, 2, 4. Layer0 is just an ordinary $ 3\times3 $ convolution. $*$ Layer5 is only required when the teacher network is VGG19 of BL \cite{ma2019bayesian}.}
		\begin{tabular}{cccccccc}
			\hline
			\multirow{2}{*}{Parameter} & \multirow{2}{*}{Expansion} & \multirow{2}{*}{No. Blocks} & \multirow{2}{*}{Stride} & \multirow{2}{*}{Kernel size} & \multicolumn{3}{c}{No. channels} \\
			\cline{6-8}
			& & & & &$1$-StuNet & $\frac{1}{2}$-StuNet  & $\frac{1}{4}$-StuNet \\
			\hline
			Layer0 &-&1&1&512& 64 &  32  &  16 \\
			Layer1 &1&1&1&512& 64 &  32  &  16 \\
			Layer2 &6&2&2&256& 128 &  64  &  32 \\
			Layer3 &6&3&2&128& 256 &  128  & 64  \\
			Layer4 &6&4&2&64& 512 &  256  &  128 \\
			*Layer5&6&4&2&32& 512 &  256  &  128 \\
			\hline
		\end{tabular}
		\label{tab:studentnet}
	\end{table*}

\subsection{Architecture of ReviewKD}
\label{sc:architecture-ReviewKD}
Due to the capacity gap between the student and teacher networks in KD models, the performance of the student network is generally limited by the teacher network. To overcome such limitation, we introduce a review mechanism following existing KD models as depicted in Figure \ref{fig:model-architecuture}, through mimicking the review mechanism of human-beings shown in Figure \ref{fig:review-mechanism}.
From Figure \ref{fig:model-architecuture}, there are an instruction phase and a review phase. In the instruction phase, a well-trained heavy teacher network in a KD model transfers its latent feature to the student network, then based on the learned feature, the student network yields an estimate of the density map by the decoder. In the review phase, similar to the simplified review mechanism of human-beings presented in Figure \ref{fig:review-mechanism}, in the first round of review, the student network uses the estimated density map as certain an attention map to pay more attention to the important crowd regions, then aggregates the interested feature yielded by the attention map and the original feature learned from the teacher network to obtain the refined feature, and finally produces a better estimate of the density map by the decoder based on such refined feature. By repeating such review procedure several rounds, the student network would yield a more and more accurate estimate of the density map. Thus, by the use of such review mechanism, the performance of the student network can be improved much and even better than that of the associated teacher network as shown by the later experiments in Section \ref{sc:experiment}.  

In the instruction phase of ReviewKD, besides the existing SKT model \cite{liu2020efficient}, we introduce a variant of SKT (VSKT) model as the KD model, in the purpose of alleviating both capacity gap and mistake propagation issues suffered by SKT as shown in Figure \ref{fig:mistake-propagation}. 
The architecture of the suggested VSKT model is depicted in Figure \ref{fig:vskt-model}.
From Figure \ref{fig:vskt-model}, similar to the SKT model, there are a well-trained teacher network, a lightweight student network, the Intra-PT and Inter-RT modules lying between the student and teacher networks to fully distill the knowledge of the teacher network, and a decoder to yield an estimate of the density map in the VSKT model. 
Here, we take CSRNet \cite{li2018csrnet} as the teacher network for example, where the heavy backbone network VGG16 is used. Besides CSRNet, some other heavy teacher networks (e.g., BL \cite{ma2019bayesian}) can be also adopted as the teacher network in the considered KD models, as shown in the later Section \ref{sc:experiment}.
We use the same decoder in the student network as in SKT \cite{liu2020efficient}.
Compared to SKT \cite{liu2020efficient}, there are two major modifications made in the suggested VSKT model.
The first one is that in the VSKT model, a new lightweight student network is adopted through taking a part of channels of the well-known MobileNetV2, instead of the teacher network as done in SKT \cite{liu2020efficient}, mainly due to the superiority of MobileNetV2 on capturing high-level features and its lightweight nature. 
Similar to SKT \cite{liu2020efficient}, the student network is also set to be a subnet of MobileNetV2 with only $1/n$ channels, where $n$ is some positive integer, say $2$ and 4 as presented in Table \ref{tab:studentnet}.
Henceforth, we call the student network with only $1/n$ channels the \textbf{$\frac{1}{n}$-StuNet}. 
The specific structure of the encoder in the student network is also presented in Table \ref{tab:studentnet}, where each layer is some residual modules from MobileNetV2.  
The second one is that VSKT only keeps the intermediate feature extraction layers while gets rid of the final decoder layer of the teacher network. Thus, distinguished to SKT, the soft ground truth as the output of the decoder layer of the teacher network is not used for the final inference. By doing this, errors made by the teacher network at the decoding stage would be avoided in VKST during the final inference stage.

\subsection{Training loss of proposed model}
In the following, we describe the training loss of the proposed ReviewKD, where we take the VSKT as an example.
As depicted in Figure \ref{fig:vskt-model}, given a scene image, we simultaneously feed it into the teacher network and $\frac{1}{n}$-StuNet for feature extraction, where the teacher network has been pre-trained on standard benchmarks.
Let Conv$i\_j$ and Layer$i\_j$ be the $j$-th convolution at the $i$-th layer of teacher and student networks, respectively.
We denote $t_i^j$ and $s_i^j$ as the features of Conv$i\_j$ and Layer$i\_j$ respectively.
Then we adopt both Intra-PT and Inter-RT modules to fully distill the teacher network. Since the Inter-RT calculates feature relations densely, we only perform distillation on some representative features, in the purpose of reducing the computational cost during the training procedure.
In the VSKT model, we use the feature extracted by the final convolution at each layer, instead of the first convolution as done in SKT \cite{liu2020efficient} for both teacher and student networks, mainly because the final convolution at each layer would capture more knowledge than the associated first convolution. Thus, the selected features are presented as follows
\begin{equation}
\label{eq:feature}
T=\left\{ t_1^1,t_1^2,t_2^2,t_3^3,t_4^3 \right\},\ S= \left\{ s_0^1,s_1^1,s_2^2,s_3^3,s_4^4 \right\}.
\end{equation}

Note that the $k$-th component of $T$ (denoted as $T_k$) and the associated component of $S$ (denoted as $S_k$) in \eqref{eq:feature} have the same resolution but different number of channels, where the number of channels of $S_k$ is only $ 1/n $ of that of $T_k$ for $k$ varying from 1 to 5.
In order to address this issue, we feed each $S_k$ into a $1\times 1$ convolution layer to yield an interim feature $\hat{S}_k$, which has the same channels to $T_k$. Thus,
$\hat{S} = \left\{ \hat{s}_0^1,\hat{s}_1^1,\hat{s}_2^2,\hat{s}_3^3,\hat{s}_4^4 \right\}.$
Based on $\hat{S}$ and following the previous work \cite[Sec. 3.2]{liu2020efficient}, the loss associated to the Intra-PT module can be expressed as
\begin{align}
\label{eq:intra-loss}
{\cal L}_{\mathrm{Intra-PT}} = \sum_{k\in |T|} \sum_{x=1}^{H_k}\sum_{y=1}^{W_k} (1-{\cal S}_k(x,y)),
\end{align}
where $|T|$ is the index set of $T$, i.e., $|T| = \{1,2,3,4,5\}$ in this case, $H_k$ and $W_k$ are respectively the height and width of feature $T_k$, and ${\cal S}_k(x,y)$ represents the similarity between $T_k$ and $\hat{S}_k$ at location $(x,y)$ defined as the following
\begin{align*}
{\cal S}_k(x,y) &= \cos(T_k(x,y),\hat{S}_k(x,y)) \\
&= \frac{1}{\ell_{k,(x,y)}^2}\sum_{c=1}^{C_k} T_k(x,y,c)\cdot \hat{S}_k(x,y,c),
\end{align*}
where $C_k$ denotes the channel number of $T_k$, $\ell_{k,(x,y)}$ denotes the length of $T_k(x,y)$, and $T_k(x,y,c)$, $\hat{S}(x,y,c)$ are respectively the response values of $T_k$ and $\hat{S}_k$ at location $(x,y)$ of the $c$-th channel. Here, we use \textit{cosine} function as the similarity measure.

Following \cite[Sec. 3.3]{liu2020efficient}, the flow of solution procedure (FSP) in the Inter-RT module can be modeled with the relationship between features from two layers. Specifically, given two features $f_1 \in \mathbb{R}^{h\times w\times m}$ and $f_2 \in \mathbb{R}^{h\times w\times n}$, the $(p,q)$-th component of their FSP matrix ${\cal F}(f_1,f_2) \in \mathbb{R}^{m\times n}$ can be defined as
\begin{align*}
 {\cal F}_{p,q}(f_1,f_2) = \sum_{x=1}^h\sum_{y=1}^w \frac{f_1(x,y,p)\cdot f_2(x,y,q)}{h\cdot w}.
\end{align*}
Notice that FSP matrix computation is implemented on features with the same resolution. However, the features in $T$ have various resolutions. To tackle this issue and simultaneously reduce the FSP computational cost, we consistently resize all features in both $T$ and $\hat{S}$ to the resolution of $t_4^3$. The resized features of $T_k$ and $\hat{S}_k$ are denoted respectively as ${\cal R}(T_k)$ and ${\cal R}(\hat{S}_k)$. In order to better capture the long-short term evolution of features, a dense FSP strategy is adopted in the proposed model, motivated by \cite{liu2020efficient}. Specifically, we generate a FSP matrix ${\cal F}({\cal R}(T_{k_1}), {\cal R}(T_{k_2}))$ for every pair $(T_{k_1},T_{k_2})$ in $T$ with $k_1 \neq k_2$, and similarly, a FSP matrix ${\cal F}({\cal R}(\hat{S}_{k_1}), {\cal R}(\hat{S}_{k_2}))$ for every pair $(\hat{S}_{k_1},\hat{S}_{k_2})$ in $\hat{S}$. Thus, the loss for the Inter-RT module can be formulated as
\begin{align}
\label{eq:inter-loss}
{\cal L}_{\mathrm{Inter-RT}} = \sum_{(k_1,k_2) \in |T|\times |T|} &\|{\cal F}({\cal R}(T_{k_1}), {\cal R}(T_{k_2})) \\
&- {\cal F}({\cal R}(\hat{S}_{k_1}), {\cal R}(\hat{S}_{k_2}))\|^2\nonumber
\end{align}

After the instruction phase, the student network yields an estimate $M_0$ of the density map via a decoder and based on the knowledge $s_4^4$ learned from the teacher network. In order to further boost performance of the student network, we stack $p$ rounds of review following the student network, as shown in Figure \ref{fig:model-architecuture}. According to the review mechanism in ReviewKD, the density map $M_i$ estimated at the $i$-th round of review can be expressed as follows
\begin{align*}
M_i = {\cal D}((s_4^4 \odot M_{i-1}) + s_4^4), \ i=1,\ldots, p.
\end{align*}
To regularize each review module in the review phase, we consider the total $L_2$ loss between the estimate at each round of review and the hard ground truth $G$, instead of only the $L_2$ loss between the eventual estimate $M_p$ of the density map after $p$ rounds of review and the hard ground truth $G$. Thus, the loss in the review phase together with the $L_2$ loss between $M_0$ and $G$ can be expressed as follows
\begin{align}
\label{eq:review-loss}
{\cal L}_{\mathrm{review}} = \sum_{i=0}^p \|M_i-G\|_F^2,
\end{align}
where $\|\cdot\|_F$ represents the Frobenius norm.
Based on the above loss, together with the loss functions \eqref{eq:intra-loss} and \eqref{eq:inter-loss} for the Intra-PT module and Inter-RT module, the training loss for the proposed model is shown as follows:
\begin{align}
\label{eq:total-loss}
&{\cal L}_{\mathrm{ReveiwKD}} = {\cal L}_{\mathrm{review}} + \lambda_1 {\cal L}_{\mathrm{Intra-PT}} + \lambda_2 {\cal L}_{\mathrm{Inter-RT}},
\end{align}
where $\lambda_1$ and $\lambda_2$ are two tunable hyper-parameters for the Intra-PT and Inter-RT loss functions.
It can be observed from \eqref{eq:total-loss} that the second and third terms are the loss introduced in the instruction phase, which aims to distill the knowledge of the teacher network to the student network, while the first term is the loss introduced in the review phase, which aims to boost the performance of the student network.

\section{Experiments}
\label{sc:experiment}

In this section, we conduct a series of experiments to demonstrate the effectiveness of the proposed model. We at first describe the experimental settings, then provide some experiments to demonstrate the effectiveness of the introduced review mechanism for reducing capacity gap, and later show the effectiveness of the proposed model via comparing with the state-of-the-art crowd counting models, and then generalize the idea of review mechanism to some heavy crowd counting models to further boost their performance, and finally present some ablation studies to determine some hyperparameters involved in the proposed model including the rounds of review in the review phase and the channel preserving ratio (CPR) used in the student network, as well as to show the superiority of the only use of hard ground truth as the supervisory information over the use of both hard and soft ground truths in the literature (e.g., \cite{liu2020efficient}).

\subsection{Experimental settings}
\label{sc:exp-setting}
In the following, we describe the experimental settings.

\textbf{A. Datasets.}
To evaluate the performance of the proposed model, we consider six benchmark datasets including ShanghaiTech Part\_A and Part\_B \cite{zhang2016single}, UCF\_CC\_50 \cite{idrees2013multi}, UCF-QNRF \cite{idrees2018composition}, WorldExpo'10 \cite{zhang2015cross} and TRANCOS \cite{guerrero2015extremely}, described as the follows:
\begin{enumerate}
\item[$\bullet$]
The \textbf{ShanghaiTech} dataset~\cite{zhang2016single} consists of Part\_A and Part\_B, where Part\_A contains 482 images and Part\_B contains 716 images. Specifically, in Part\_A, there are 300 images for training and 182 images for testing, while in Part\_B, there are 400 images for training and 316 images for testing.
	
\item[$\bullet$]
The \textbf{UCF\_QNRF} dataset is a large-scale crowd dataset recently released in \cite{idrees2018composition}. It contains 1,535 images in which there are about 1.25 million head annotations. There are 1,201 images for training and 334 images for testing.

\item[$\bullet$]
The \textbf{UCF\_CC\_50} dataset~\cite{idrees2013multi} contains 50 images with a total of 63,974 head annotations. The number of heads per image varies greatly in a wide range from 94 to 4,543.
For the fairness of comparison, we use the standard settings as done in \cite{idrees2013multi}.

\item[$\bullet$]
The \textbf{WorldExpo'10} dataset~\cite{zhang2015cross} is a large scale data-driven cross-scene crowd counting dataset collected from Shanghai 2010 WorldExpo.
There are 3,380 frames of images for training and 600 frames of images for testing, where the test set is divided into five test sets, denoted as \textit{S1, S2, S3, S4} and \textit{S5}.

\item[$\bullet$]
The \textbf{TRANCOS} dataset~\cite{guerrero2015extremely} is a dataset constructed for vehicle counting in traffic jam scenes. It contains 1,244 traffic images with vehicle numbers varying from 9 to 107. There are a total of 46,796 annotated vehicles with extreme
overlap.

\end{enumerate}
	
\textbf{B. Evaluation metrics.}
For most of benchmark datasets, we use the commonly used mean absolute error (MAE) and mean square error (MSE) as evaluation metrics to evaluate the performance of models. Specifically, MAE and MSE are defined as follows:
\begin{equation*}
\text{MAE}=\frac{1}{N} \sum _{i=1}^N\vert C_i-C_i^{\mathrm{tr}}\vert, \ \text{MSE} = \sqrt{\frac{1}{N} \sum _{i=1}^N\vert C_i-C_i^{\mathrm{tr}}\vert^2},
\end{equation*}
where $C_i$ and $C_i^{\mathrm{tr}}$ respectively represent the predicted counting number and ground truth.

For the TRANCOS dataset, motivated by the existing literature \cite{guerrero2015extremely}, we use the Grid Average Mean absolute Error (GAME)  instead of MAE and MSE as the evaluation metric, where GAME is defined as follows: 
\begin{equation}
\label{gamemetric}
\small
	\text{GAME}^L = \frac{1}{N} \sum _{i} \sum _{l=1}^{2^L}\vert C_i^l-\hat{C_i^l}\vert,
\end{equation}
where $\text{GAME}^L$ evaluates the case that images are divided into $2^L$ non-overlapping regions. 
For all these evaluation metrics, the smaller value implies the better performance.

\textbf{C. Implementation detail.}
We use the popular optimizer \textit{Adam} \cite{kingma2014adam} to train the proposed model, where the learning rate and the parameter of weight decay used in Adam are empirically set as 0.0001 and 0.0005, respectively, and the batch size is set to be 1.
The hyperparameters $\lambda_1$ and $\lambda_2$ in \eqref{eq:total-loss} are empirically set to be 0.5.
Experiments are implemented on a single NVIDIA 3090 GPU.
Moreover, the same data augmentation method used in the training of teacher network is also adopted to train the proposed model. 
The codes are available at \url{https://github.com/Dearbreeze/ReviewKD}.

\begin{table*}[!h]
		\setlength{\tabcolsep}{1.4mm}
		\centering
		\caption{The performance of the proposed models over the ShanghaiTech Part\_A,  Part\_B, UCF\_CC\_50 and UCF\_QNRF datasets with comparison to the teacher networks. For each block, the first row presents the performance of the teacher network, while the second and third rows present the performance of the proposed ReviewKD-SKT and ReviewKD-VSKT, respectively. The best results are highlighted in bold. $*$ These teacher networks are implemented by ourselves.}
		\begin{tabular}{l|ccc|ccc|ccc|ccc|c}
		\hline	
			\multirow{2}{*}{Model} & \multicolumn{3}{c|}{ShanghaiTech Part\_A} & \multicolumn{3}{c|}{ShanghaiTech Part\_B} & \multicolumn{3}{c|}{ UCF\_CC\_50} & \multicolumn{3}{c|}{UCF\_QNRF} &	\multirow{2}{*}{Params} \\
			\cline{2-13}
			& MAE  & MSE & $\mathrm{Ratio}$  & MAE & MSE   & $\mathrm{Ratio}$   & MAE   & MSE  &$\mathrm{Ratio}$ & MAE   & MSE  &$\mathrm{Ratio}$ \\
			\hline
			CSRNet* & 68.31&109.28& & 10.23&16.50 & & 263.43 & 388.33 & &190.21&381.34&  &16.26M\\
			ReviewKD-SKT &67.83  &109.30 &1.00 & 9.81 & 16.17 &1.03 &247.51  &352.24  &1.09 & 137.39& 207.73& 1.66&1.02M \\
			ReviewKD-VSKT &\textbf{67.33} & \textbf{106.87} & \textbf{1.02}&\textbf{8.99}&\textbf{13.39}& \textbf{1.19} & \textbf{214.71}&\textbf{293.28}& \textbf{1.28}  &\textbf{104.76}&\textbf{179.63}&\textbf{2.01} &1.38M\\
			\hline
			BL* & 68.50&110.96& & 7.65&12.63 & &227.56&313.44& &89.20&156.66&  &21.50M \\
			ReviewKD-SKT & 62.22& 103.78& 1.08& 7.33&12.51  &1.02 &211.54 &300.67 &1.06 &89.37 &153.55 &1.01 &1.35M\\
			ReviewKD-VSKT &\textbf{61.47} &\textbf{ 101.26} & \textbf{1.10}&\textbf{7.16}&\textbf{11.95}& \textbf{1.06}& \textbf{191.97}&\textbf{277.63}& \textbf{1.15} &\textbf{88.22}&\textbf{148.99}&\textbf{1.04} &2.20M\\
		\hline 
		\end{tabular}		
		\label{tab:beyond-shanghai}
	\end{table*}

\begin{table*}
		\centering
		\setlength{\tabcolsep}{1.4mm}
		\caption{The performance of the proposed models over the WorldExpo'10 and TRANCOS datasets with comparison to the teacher networks. For WorldExpo'10 dataset, only MAE is recorded for five different test sets, while for the TRANCOS dataset, the GAME metrics \eqref{gamemetric} with four different $L$ are recorded. \textit{Avg.} represents the average of evaluation metrics.}
		\begin{tabular}{l|ccccccc|cccccc}
			\hline 
			\multirow{2}{*}{Model} & \multicolumn{7}{c|}{WorldExpo'10} & \multicolumn{6}{c}{TRANCOS}\\
			\cline{2-14}
			& S1 & S2 & S3 & S4 & S5 & Avg. & $\mathrm{Ratio}$ & GAME$^0$ & GAME$^1$ & GAME$^2$ & GAME$^3$ & Avg. & $\mathrm{Ratio}$\\
			\hline
			CSRNet* & 2.0&14.5&10.6&12.7&3.3&8.6& & 3.53&5.63&8.72&13.80 &7.92 & \\
			ReviewKD-SKT & 1.7 &12.8 &9.5 &12.1 &3.7 &8.0 & 1.08& 3.23&5.70 &7.93 &12.31  &7.29& 1.09\\
			ReviewKD-VSKT &\textbf{1.5}&\textbf{10.9}&\textbf{7.9}&\textbf{12.0}&\textbf{2.7}&\textbf{7.0}&\textbf{1.23} &\textbf{3.24}&\textbf{4.93}&\textbf{7.34}&\textbf{11.47} &{\bf 6.75} &\textbf{1.17}\\
			\hline	
			BL* & 1.8&13.9&8.2&10.8&2.8&7.5& & 3.43&4.78&7.01&11.32 &6.64&\\
			ReviewKD-SKT & 1.9&11.7 &8.4 &10.1 &2.6 &6.9 &1.09 &3.28  &4.77 &6.85 &10.22  &6.27 &1.06\\
			ReviewKD-VSKT &\textbf{1.7}&\textbf{10.9}&\textbf{7.8}&\textbf{9.6}&\textbf{2.3}&\textbf{6.5}&\textbf{1.15} &\textbf{3.08}&\textbf{4.22}&\textbf{6.13}&\textbf{9.97} & {\bf 5.85} &\textbf{1.14}\\
			\hline 
		\end{tabular}
		\label{tab:beyond-worldexpo-trancos}
	\end{table*}

\subsection{Reducing capacity gap by review mechanism}
\label{sc:effectiveness-review}
In this subsection, we conduct a series of experiments over six benchmark datasets described in Section \ref{sc:exp-setting} to demonstrate the effectiveness of the introduced review mechanism for alleviating the capacity gap issue, through showing that the proposed ReviewKD can yield the performance beyond the associated teacher network. 
In the instruction phase, we consider two KD models, i.e., SKT and VSKT suggested in this paper. Thus, we denote the proposed ReviewKD with SKT and VSKT as \textit{ReviewKD-SKT} and \textit{ReviewKD-VSKT}, respectively. 
Besides CSRNet, we also consider another representative heavy teacher network, i.e., BL \cite{ma2019bayesian} in the instruction phase. When taking BL as the teacher network, the student network in the suggested VSKT model should be slightly modified with one more layer, as presented in Table \ref{tab:studentnet}. 
As demonstrated by the later experiments in Section \ref{sc:rounds-review}, we use two rounds of review in the review phase.

\begin{figure*}[!t]
\begin{minipage}[b]{1\linewidth}
\centering
\includegraphics*[scale=.58]{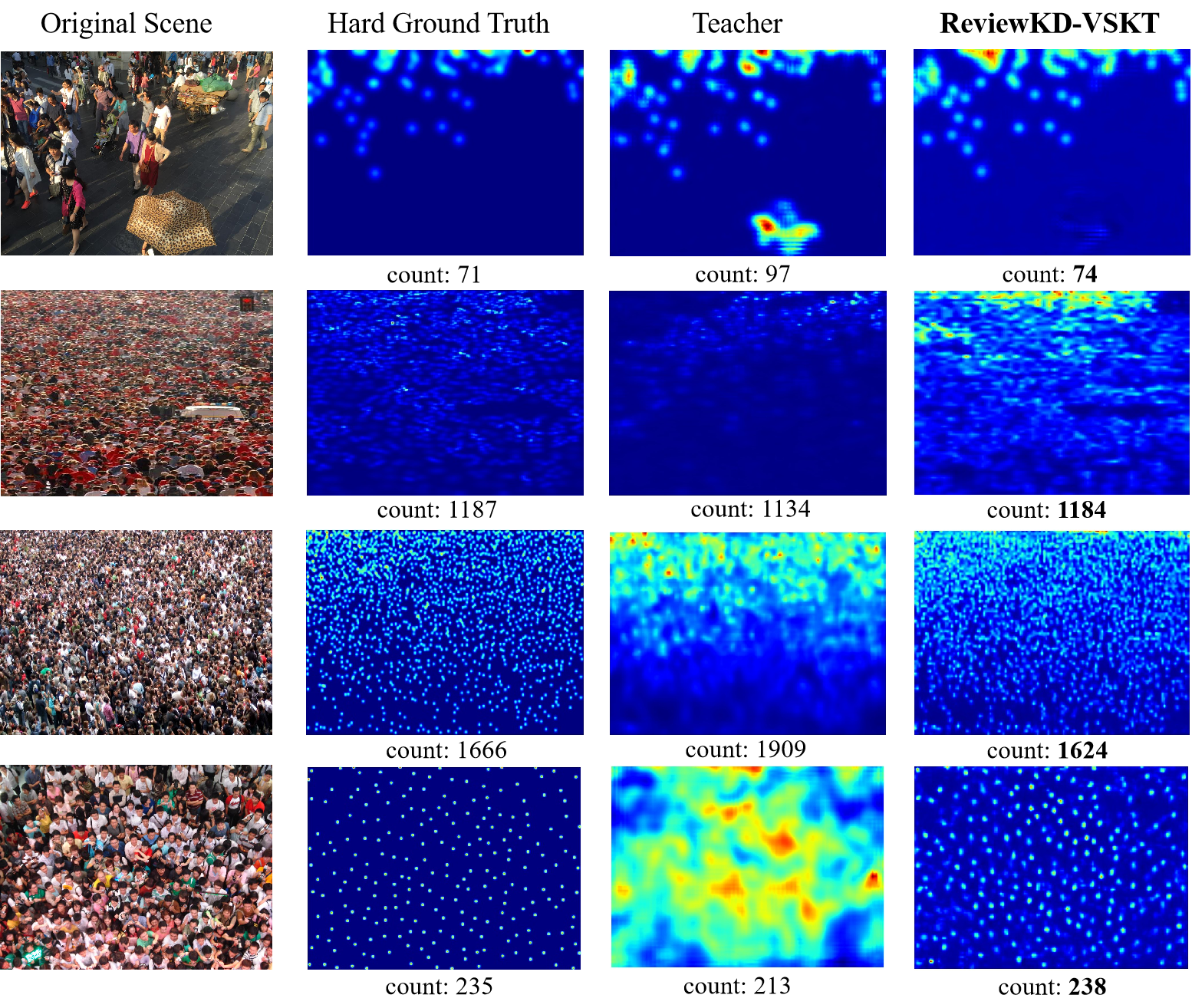}
\end{minipage}
\hfill
\caption{Some visualization results of the performance of the proposed ReviewKD-VSKT model. The first scene is from ShanghaiTech\_Part B dataset, and the second to the fourth scenes are from UCF\_QNRF datasets. It can be observed that the estimates of the crowd counts yielded by the proposed model are much more accurate than the associated teacher network. 
}
\label{fig:beyond-examples}
\end{figure*}

In order to quantify the boosting ratio of the proposed model as compared to the associated teacher network, we define the boosting ratio as the following:
\begin{align*}
\mathrm{Ratio} = \frac{\mathrm{Performance \ of\ teacher}}{\mathrm{Performance\ of\ student}},
\end{align*}
where the {\bf Performance of teacher (respectively, student)} is defined as MAE+MSE of the teacher network (respectively the student network) for the ShanghaiTech Part\_A, Part\_B, UCF\_CC\_50 and UCF\_QNRF, and MAE for the WorldExpo'10 dataset, and GAME$^L$ for the TRANCOS dataset, respectively.
When $\mathrm{Ratio}$ is above one, it implies that the performance of the proposed ReviewKD model is beyond that of the teacher network, and a larger $\mathrm{Ratio}$ implies the better boosting performance of the proposed model. The experiment results are presented in Tables \ref{tab:beyond-shanghai} and \ref{tab:beyond-worldexpo-trancos}. Some visualization results of the proposed model are presented in Figure \ref{fig:beyond-examples}.

From Tables \ref{tab:beyond-shanghai} and \ref{tab:beyond-worldexpo-trancos}, the performance of the proposed models is better than the associated teacher network over these six benchmark datasets, in terms of the concerned evaluation metrics, as well as the boosting ratio. 
When comparing the performance of the proposed models using different teacher networks, the proposed models using BL model as the teacher network generally outperform the proposed models using the CSRNet as the teacher network, since the performance of BL model is generally better than that of CSRNet. This implies that a better teacher network generally leads to a better student network. When concerning the boosting ratio, we can observe that the proposed models using CSRNet as the teacher network yield larger improved ratios than the associated counterparts using BL as the teacher network. This demonstrates that the introduced review mechanism will bring more improvement when using a worse teacher network. Moreover, when comparing the performance of ReviewKD-SKT and ReviewKD-VSKT,  it can be observed from Table \ref{tab:beyond-shanghai} and Table \ref{tab:beyond-worldexpo-trancos} that the performance of ReviewKD-VSKT is better than that of ReviewKD-SKT. This shows that the suggested VSKT is generally better than SKT \cite{liu2020efficient}.
From Figure \ref{fig:beyond-examples}, we can observe that the estimates of crowd counts yielded by the proposed ReviewKD-VSKT model are more accurate than those of the associated teacher networks. These experiment results demonstrate that the introduced review mechanism can effectively alleviate the capacity gap issue suffered by existing KD models and in particular break through the limitation of the performance of the associated teacher network. Besides, the proposed models have much smaller sizes of parameters than the associated teacher networks, as shown in the last column of Table \ref{tab:beyond-shanghai},  where the parameter sizes of the proposed ReviewKD models are only about ten percentages of those of the teacher networks.

\begin{table*}[t]
	\setlength{\tabcolsep}{0.8mm}
	\centering
	\caption{Comparison results of different lightweight models on five benchmark datasets including the ShanghaiTech Part\_A, ShanghaiTech Part\_B, UCF\_QNRF, UCF\_CC\_50 and WorldExpo'10 datasets. The best results are highlighted in bold. There are five different test sets for the WorldExpo'10 dataset and only MAE are record for this dataset.}
	\begin{tabular}{l|cc|cc|cc|cc|cccccc|cc}
		\hline 
		\multirow{2}{*}{Model} & \multicolumn{2}{c|}{Part\_A} & \multicolumn{2}{c|}{Part\_B} & \multicolumn{2}{c|}{UCF\_QNRF} & \multicolumn{2}{c|}{UCF\_CC\_50} & \multicolumn{6}{|c|}{MAE on WorldExpo'10} &	\multirow{2}{*}{Params} \\
		\cline{2-15}
		& MAE & MSE  & MAE & MSE & MAE & MSE & MAE & MSE & S1 & S2 & S3 & S4 & S5 & Avg. &\\
		\hline
		SANet \cite{cao2018scale}                          &67.0 & 104.5 &8.4&13.6&-&- & 254.8 & 334.9 & 2.6 & 13.2 &9.0& 13.3 & 3.0 & 8.2 &0.91M\\
		ACSCP \cite{shen2018crowd}                         &75.7 &102.7 & 17.2&27.4 & -&- & 291.0 & 404.6& 2.8 & 14.1 & 9.6 & 8.1 & 2.9 & 7.5 &5.10M\\
		TEDNet \cite{jiang2019crowd}      & 64.2 & 109.1  & 8.2  & 12.8 & 113.0 & 188.0  &249.4  & 354.5  & 2.3 & \textbf{10.1} & 11.3 & 13.8 & 2.6 & 8.0   &1.63M\\
		{\small MobileCount \cite{wang2020mobilecount}} &84.4&135.1&8.6&13.8&127.7&216.5 &284.8& 392.8 &-& -&- & -&- &11.1 &3.40M\\
		SKT \cite{liu2020efficient} & 62.73 & 102.33 &7.98 & 13.13 & 96.24 & 156.82 &235.86 &322.33 &\textbf{1.4}&10.5&13.1&\textbf{7.6}&4.1&7.3 &1.35M\\
		ReviewKD-SKT & 62.22& 103.78& 7.33&12.51  &89.37 &153.55 &211.54 &300.67 & 1.9&11.7 &8.4 &10.1 &2.6 &6.9   &1.35M\\
		VSKT &63.72 & 105.39& 7.74& 13.07 &92.04&155.52&228.11&309.74&2.1 & 12.3 & 8.6 & 10.4 & 2.6 & 7.2  &2.20M\\
		ReviewKD-VSKT &\textbf{61.47} & \textbf{101.26} &\textbf{7.16}&\textbf{11.95}  &\textbf{88.22}&\textbf{148.99} &\textbf{191.97}&\textbf{277.63}&1.7&10.9&\textbf{7.8}&9.6&\textbf{2.3}&\textbf{6.5} &2.20M\\
		\hline 
	\end{tabular}
	\label{tab:comp-sota}
\end{table*}

\begin{table*}
		\centering
	\caption{Comparison results of different models on the TRANCOS dataset. The best results are highlighted in bold.}
		\begin{tabular}{l|cccc|c}
			\hline 
			Model & GAME$^0$ & GAME$^1$ & GAME$^2$ & GAME$^3$ & Params \\
			\hline
			CSRNet \cite{li2018csrnet} & 3.67 & 5.82 & 8.53 & 14.40 & 16.26M\\
			
			COBC \cite{liu2019counting}   & 3.15 & 5.45 & 8.34 & 15.02 & 14.7M\\
			
			DensityCNN-H \cite{jiang2020density} & 3.17 &4.78 &6.30 &\textbf{8.26}& $\geq$10M\\
            
            LSC-CNN \cite{sam2020locate} &4.6&5.4&6.9&8.3& 20.6M\\
            
            KDMG \cite{wan2020kernel}  & 3.13 & 4.79 & 6.20 & 8.68 & $\geq$10M \\
            
            SKT  &  3.45 & 5.12  &7.13  & 10.58  & {\bf 1.35M} \\
            ReviewKD-SKT &3.28  &4.77 &6.85 &10.22 &{\bf 1.35M} \\
            VSKT  & 3.41  & 4.82  &7.15  & 10.40  & {\bf 2.20M} \\
            
			ReviewKD-VSKT &\textbf{3.08}&\textbf{4.22}&\textbf{6.13}&9.97&{\bf 2.20M}\\
			\hline 
		\end{tabular}
	\label{tab:comp-TRANCOS}
\end{table*}

\subsection{Comparison with state-of-the-art models}
In this subsection, we compare the performance of the proposed model (using VSKT as the KD model and BL \cite{ma2019bayesian} as the teacher network) with the state-of-the-art lightweight crowd counting models  including SANet \cite{cao2018scale}, ACSCP \cite{shen2018crowd}, TEDNet \cite{jiang2019crowd}, MobileCount \cite{wang2020mobilecount}, SKT \cite{liu2020efficient} and VSKT suggested in this paper, where both SKT and VSKT also use the BL as the teacher network. Moreover, since there are very few lightweight crowd counting models evaluated on the TRANCOS dataset, we compare the performance of the proposed model with the state-of-the-art heavy crowd counting models including CSRNet \cite{li2018csrnet}, COBC \cite{liu2019counting}, DensityCNN-H \cite{jiang2020density}, LSC-CNN \cite{sam2020locate} and KDMG \cite{wan2020kernel} over the TRANCOS dataset.
The comparison results are presented in Table \ref{tab:comp-sota} and Table \ref{tab:comp-TRANCOS}. 

From Table \ref{tab:comp-sota},
the proposed model outperforms the state-of-the-art lightweight crowd counting models over ShanghaiTech Part\_A and Part\_B, UCF\_CC\_50 and UCF\_QNRF datasets in terms of both MAE and MSE, and achieves the best performance in average over the  WorldExpo'10 dataset, where the proposed model achieve the best performance on the S3 and S5 test sets. From Table \ref{tab:comp-TRANCOS}, the proposed model achieves the best performance on the TRANCOS dataset in terms of GAME$^0$, GAME$^1$ and GAME$^2$ in comparison to the state-of-the-art heavy crowd counting models, while the size of model parameters of the proposed ReviewKD is much smaller than those of heavy models, as presented in the last column of Table \ref{tab:comp-TRANCOS}. These demonstrate the effectiveness of the proposed model.

When comparing the performance of ReviewKD-VSKT and the associated KD model, i.e., VSKT presented in the last two rows of Table \ref{tab:comp-sota} and Table \ref{tab:comp-TRANCOS}, it can be observed that the performance of VSKT can be substantially improved by introducing a review mechanism. This also demonstrates the effectiveness of the introduced review mechanism. Moreover, when concerning the performance of SKT and VKST presented in the fourth and second rows from the end of Table \ref{tab:comp-sota} and Table \ref{tab:comp-TRANCOS}, we can observe that VSKT outperforms SKT over most of benchmark datasets, which implies the effectiveness of the suggested VSKT.

\subsection{Generalization to heavy crowd counting models}
\label{sc:generalization-heavy-model}
\begin{figure*}[!t]
\begin{minipage}[b]{1\linewidth}
\centering
\includegraphics*[scale=.55]{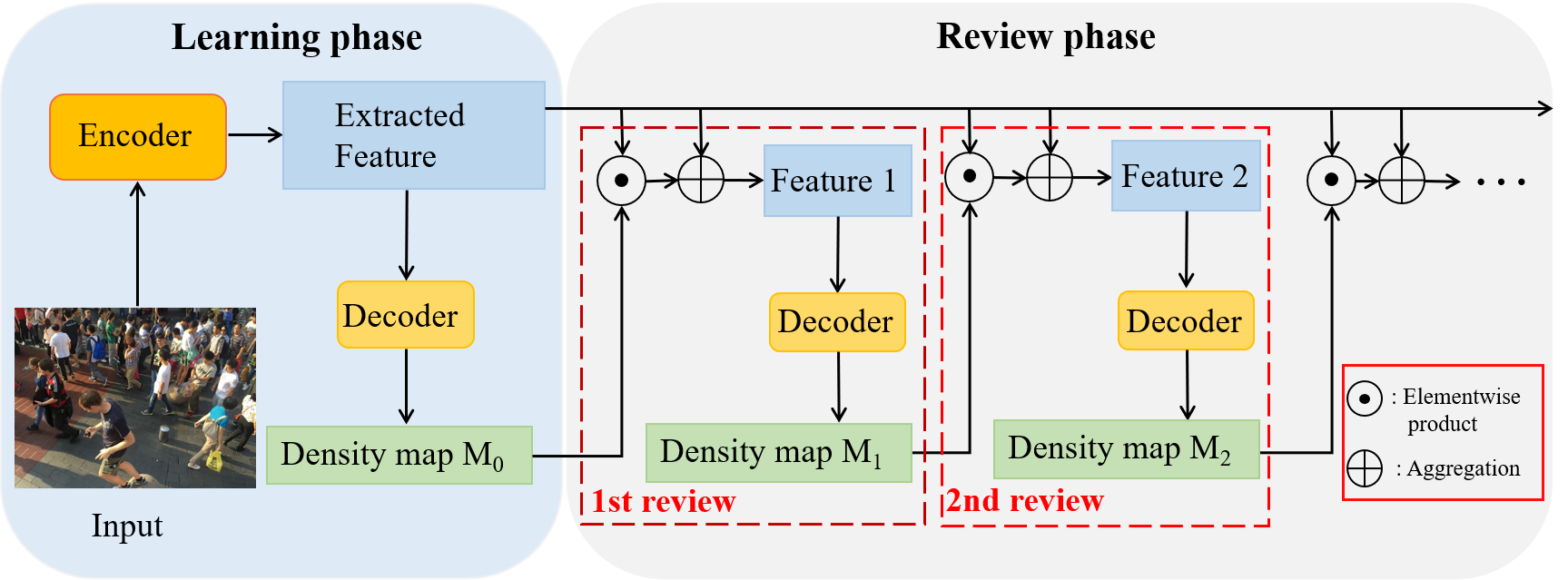}
\end{minipage}
\hfill
\caption{An overview of heavy crowd counting models equipped with review mechanism.
}
\label{fig:heavymodel-review-arch}
\end{figure*}

In this subsection, we conduct a set of experiments to show that the introduced review mechanism can be used as a plug-and-play module to boost the performance of a kind of heavy crowd counting models without adding any additional parameter. The architecture of the heavy crowd counting models equipped with the review mechanism is presented in Figure \ref{fig:heavymodel-review-arch}. From Figure \ref{fig:heavymodel-review-arch}, an encoder-decoder based heavy crowd counting model is used in the learning phase to generate the feature and an estimate $M_0$ of the density map, then a review mechanism can be applied to improve the performance of such heavy crowd counting models. Note that in the review phase, we use the same decoder of the crowd counting models, and thus do not introduce any additional model parameter.   

To demonstrate the effectiveness of the introduced review mechanism when adapted to the heavy crowd counting models, we consider three representative heavy crowd counting models, i.e., BL \cite{ma2019bayesian}, CSRNet \cite{li2018csrnet} and MCNN \cite{zhang2016single}.
Motivated by the later experiments in Section \ref{sc:rounds-review}, we also take two rounds of review in the review phase.
We compare the performance of these heavy models as well as their counterparts with two rounds of review over the ShanghaiTech Part\_B dataset. Experiment results are presented in Table \ref{tab:review-pnp}. It can be observed from Table \ref{tab:review-pnp} that the performance of these heavy crowd counting models can be substantially improved by equipping with the introduced review mechanism, and the more improvement can be achieved when adapted to a worse heavy model.

\begin{table}[h]
		\centering
		\caption{Improved performance of existing heavy crowd counting models through equipping with the review mechanism over ShanghaiTech Part\_B dataset. For each block, the first row presents the performance of the original crowd counting model, while the second row presents the performance of the associated modified model equipped with two rounds of review.}
		\begin{center}
			\begin{tabular}{l|cc|c}
			\hline 
				Model & MAE & MSE & Params\\\hline
				BL\cite{ma2019bayesian} & 7.88 & 12.77 & 21.50M\\
				BL + Review & \textbf{7.59} & \textbf{12.22} & 21.50M\\
				\hline
				CSRNet\cite{li2018csrnet} & 10.23 & 16.50 & 16.26M\\
				CSRNet + Review  & \textbf{8.63} & \textbf{14.15} & 16.26M\\
				\hline
				MCNN\cite{zhang2016single}  & 25.20 & 41.80 & 0.13M\\
				MCNN + Review  & \textbf{18.30} & \textbf{29.00} & 0.13M\\
		\hline 
			\end{tabular}

		\end{center}
		\label{tab:review-pnp}
	    \vspace{-5pt}
		\vspace{-5pt}
			\vspace{-5pt}
	\end{table}

\subsection{Ablation studies}
\label{sc:ablation}
In this subsection, we provide a set of ablation studies to show the effect of some hyperparameters including the rounds of review in the review mechanism and the channel preservation rate (CPR), i.e., $\frac{1}{n}$ in the student network of the VSKT model, as well as the superiority of the use of only hard ground truth as compared to the use of both hard and soft ground truths in the literature \cite{liu2020efficient}. 

\subsubsection{On rounds of review}
\label{sc:rounds-review}
\begin{figure}[!t]
\begin{minipage}[b]{0.49\linewidth}
\centering
\includegraphics*[scale=.31]{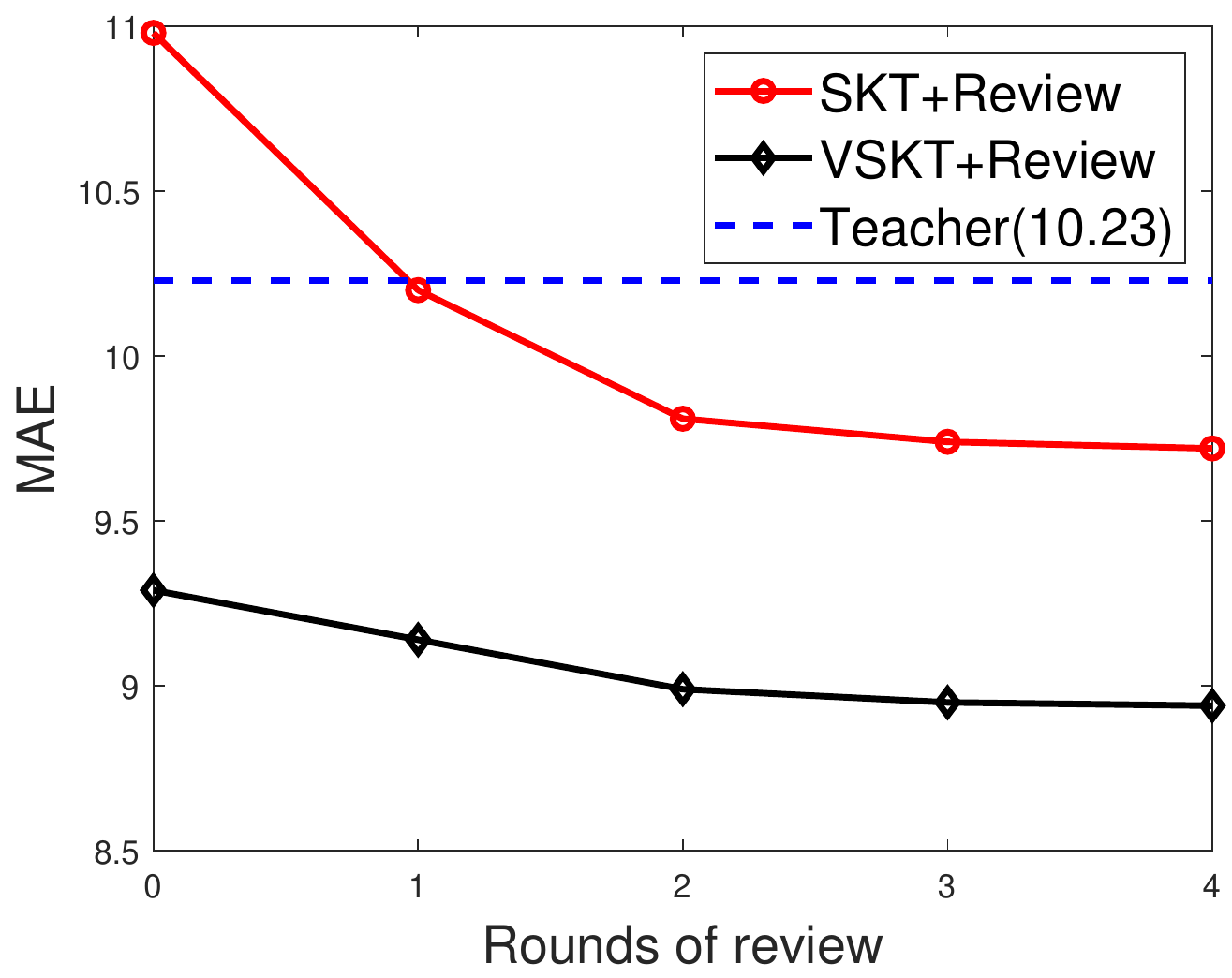}
\centerline{{\small (a) MAE}}
\end{minipage}
\hfill
\begin{minipage}[b]{0.49\linewidth}
\centering
\includegraphics*[scale=.31]{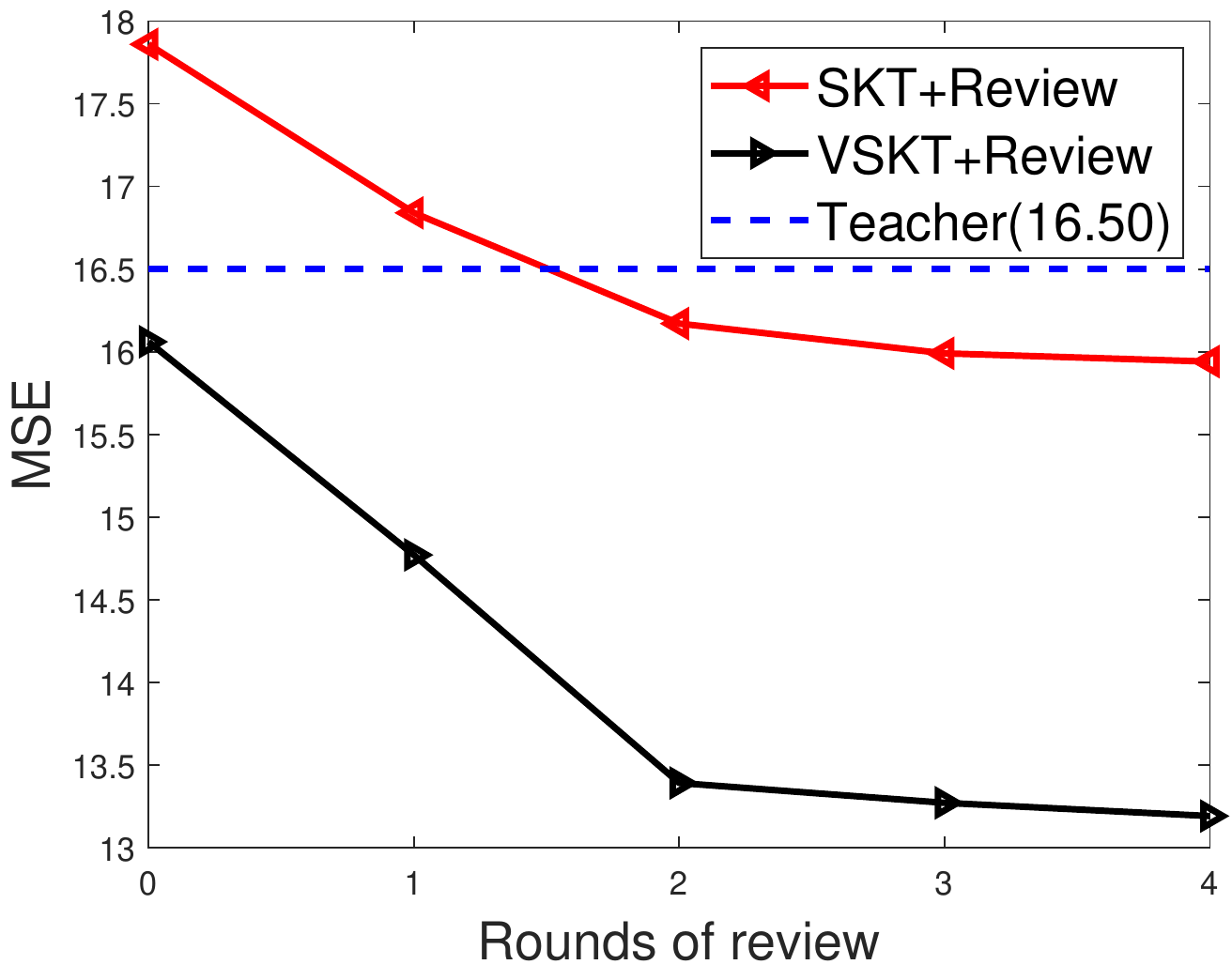}
\centerline{{\small (b) MSE}}
\end{minipage}
\hfill
\caption{The influence of the number of rounds of review in the review phase over ShanghaiTech Part\_B dataset.
}
\label{fig:review-module}
\end{figure}

As depicted in Figure \ref{fig:model-architecuture}, there is a hyperparameter, i.e., the number of rounds of review used in the review phase. 
In this subsection, we conduct a set of experiment over the ShanghaiTech Part\_B dataset to show the effect of this hyperparameter.
In these experiments, we consider two different KD models in the instruction phase, i.e., SKT \cite{liu2020efficient} and VSKT suggested in this paper, where the teacher networks used for both SKT and VSKT are CSRNet \cite{li2018csrnet}, while the student networks of SKT and VSKT are $\frac{1}{4}$-CSRNet and $\frac{1}{4}$-StuNet, respectively. We use the channel preserving ratio $\frac{1}{4}$ as the default value for both student networks in SKT and VSKT mainly motivated by \cite{liu2020efficient} and the later experiments shown in Section \ref{sc:CPR}.  For each KD model, we consider four different rounds of review in the review phase, i.e., $\{1,2,3,4\}$.  
The curves of MAE and MSE with respect to the number of rounds of review are respectively depicted in Figure \ref{fig:review-module} (a) and (b). 

From Figure \ref{fig:review-module}, it can be observed that the performance of both KD models can be substantially improved by incorporating the review mechanism in terms of both MAE and MSE. Specifically, by Figure \ref{fig:review-module}, the performance of the proposed ReviewKD models with only one round of review is much better that of the associated KD models. 
It can be also observed from Figure \ref{fig:review-module} that both ReviewKD models with two rounds of review outperform the associated teacher network, i.e., CSRNet in these experiments. This demonstrates that the introduced review mechanism is feasible to address the capacity gap issue and in particular can break through the limitation of the teacher network in KD models such that the student network can achieve the performance beyond its associated teacher network. 

When concerning the effect of the number of rounds of review, we can observe that the performance of both ReviewKD models gets better as the number of rounds of review increasing, and improves a bit when the rounds of review are no more than two, while gains less improvement when the rounds of review are above two. This shows that two rounds of review are in general sufficient in the proposed ReviewKD models for the consideration of both performance and computational cost. Thus, we suggest using two rounds of review in the proposed ReviewKD models in the later experiments. 

\begin{figure*}[!t]
\begin{minipage}[b]{1\linewidth}
\centering
\includegraphics*[scale=.5]{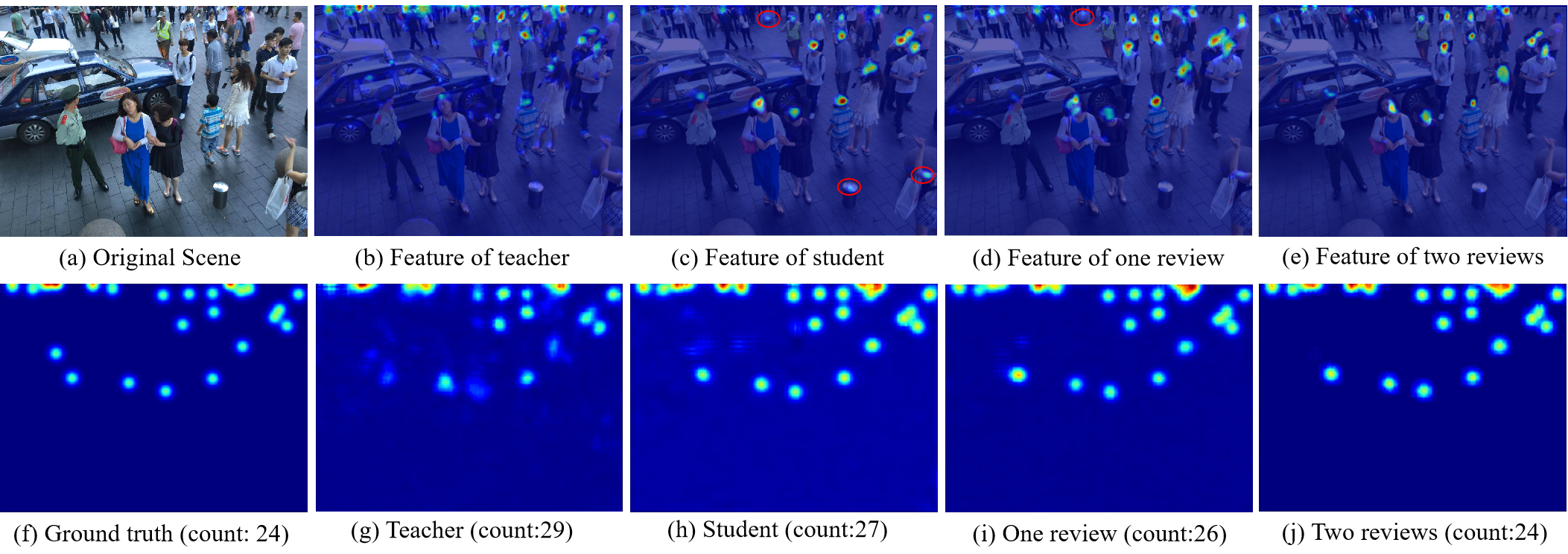}
\end{minipage}
\hfill
\caption{Some visualization results of the proposed ReviewKD-VSKT over the ShanghaiTech Part\_B dataset.
At the first row, figures (b)-(e) present the feature of the teacher network, student network, and ReviewKD-VSKT with one and two rounds of review, respectively, and figures (g)-(j) at the second row present the associated estimates of density maps as well as the crowd counts presented in the brackets. }
\label{fig:review-visualization}
\end{figure*}

We also provide some visualization results of ReviewKD with VSKT as the KD model in Figure \ref{fig:review-visualization} to show the learned features and estimated density maps during the review phase. From Figure \ref{fig:review-visualization} (c)-(e) and (h)-(j), the feature as well as the estimated density map becomes more and more accurate as the number of rounds of review increasing. Particularly, for the concerned scene image, the proposed ReviewKD model with two rounds of review yields the exact estimate of the crowd count as shown in Figure \ref{fig:review-visualization}(j). This also shows the effectiveness of the proposed review mechanism.

When considering the performance of ReviewKD with these two different KD models, it can be observed from Figure \ref{fig:review-module} that ReviewKD-VSKT generally outperforms ReviewKD-SKT in terms of both MAE and MSE. In particular, the performance of SKT is slightly worse than the teacher network, while the performance of VSKT is better than the teacher network. This shows that the suggested lightweight student network VSKT should reduce the capacity gap between the teacher and student networks and thus yield better performance.
Moreover, it can be observed from Figure \ref{fig:mistake-propagation} that the error propagation issue suffered by SKT can be effectively alleviated by the use of the new lightweight student network in VSKT. By Figure \ref{fig:mistake-propagation}(a), there is an obscured region by an umbrella in the scene image. Due to such obscuration, the teacher network would make a mistake at this region as shown by Figure \ref{fig:mistake-propagation}(c), and propagate it to the student network in SKT as depicted in Figure \ref{fig:mistake-propagation}(d), while such mistake can be effectively avoided by the suggested VSKT as shown in Figure \ref{fig:mistake-propagation}(e), through introducing a new lightweight student network constructed from the well-known MobileNetV2 instead of the associated teacher network and only using the hard ground truth as the supervisory information.

	\begin{table}[!h]
		\centering
		\caption{Comparison performance of the proposed ReviewKD-VSKT model with \textit{hard+soft} ground truths and only the \textit{hard} ground truth over ShanghaiTech Part\_A and Part\_B dataset.}
		\begin{center}
			\begin{tabular}{l|cc|cc}
			\hline 
				\multirow{2}{*}{Model}    & \multicolumn{2}{c|}{Part\_A} & \multicolumn{2}{c}{Part\_B} \\
				\cline{2-5}
				& MAE & MSE & MAE & MSE \\
				\hline
				Hard + Soft & 68.91 & 112.11 & 10.10 & 16.36\\
			    Hard  & {\bf67.33} & {\bf106.87} & {\bf 8.99} & {\bf 13.39}  \\
            \hline 
			\end{tabular}
		\end{center}
		\label{tab:hard-vs-soft}
	\end{table}

\subsubsection{Hard vs. hard+soft ground truths}
\label{sc:hard-ground-truth}

As analyzed in Section \ref{sc:architecture-ReviewKD} and shown in Figure \ref{fig:mistake-propagation}, the soft ground truth as the output of the decoder of the teacher network may contain some errors. When the soft ground truth is used as the supervisory information for inference, these errors made by the teacher network may propagate to the student network, resulting in the performance degradation of the student network. Thus, as described in Section \ref{sc:proposed-model}, we only use the hard ground truth in the suggested VSKT model as the supervisory information for the final inference, instead of both soft and hard ground truths used in SKT \cite{liu2020efficient}. To verify this, we conduct several experiments over ShanghaiTech Part\_A and Part\_B datasets. Specifically, in these experiments, we consider the proposed ReviewKD-VSKT model (where the teacher and student networks are respectively CSRNet and $\frac{1}{4}$-StuNet presented in Table \ref{tab:studentnet} and two rounds of review are used) and its variant with both soft and hard ground truths as similarly done in SKT, that is, the decoder layer in the teacher network is also kept in the variant as the supervisory information. Experiment results are presented in Table \ref{tab:hard-vs-soft}.
From Table \ref{tab:hard-vs-soft}, the performance of the proposed model using only the hard ground truth as the supervisory information is better than that of its variant with both soft and hard ground truths as the supervisory information, mainly due to the avoidance of errors in the soft ground truth by getting rid of it at the inference stage.

\subsubsection{On channel preservation rate in student network}
\label{sc:CPR}
The choice of the channel preservation rate (CPR), i.e., $\frac{1}{n}$, plays an important role in the tradeoff between performance and scale of the student network. To determine $n$, we conduct a series of experiments over ShanghaiTech Part\_B dataset to show the influence of CPR. Specifically, we consider the proposed ReviewKD-VSKT model with two rounds of review as suggested in the previous experiments, and five different CPRs, that is, $\{1, \frac{1}{2}, \frac{1}{3}, \frac{1}{4}, \frac{1}{5}\}$ for the student network. The experiment results are presented in Table \ref{tab:CPR}. It can be observed from Table \ref{tab:CPR} that the proposed ReviewKD-VSKT with the $\frac{1}{4}$-StuNet achieves the optimal tradeoff between performance and scale of the model. As a consequence, we suggest using $\frac{1}{4}$ as the default CPR value in the proposed ReviewKD-VSKT model.

\begin{table}[!h]
	\centering
	\caption{The effect of channel preservation rate used in the student network of the proposed ReviewKD-VSKT model.}
	\begin{center}
		\setlength{\tabcolsep}{1.3mm}
		\begin{tabular}{l|ccccc}
		\hline 
			CPR ($1/n$) & 1 & $\frac{1}{2}$  & $\frac{1}{3}$  & ${\bf \frac{1}{4}}$   & $\frac{1}{5}$ \\\hline
			MAE & 8.92 & 8.94 & 8.97 & 8.99 & 9.43\\ 
            MSE & 13.11 & 13.37 & 13.66 & 13.39 & 15.71\\
            Params & 21.71 M & 5.64 M & 2.54 M & 1.49 M & 0.96 M\\
         \hline 
		\end{tabular}		
	\end{center}
	\label{tab:CPR}
\end{table}

\section{Conclusion}
\label{sc:conclusion}

The capacity gap issue suffered by KD models generally limits their performance in crowd counting. To address this issue, we suggest a novel review mechanism following the KD models, inspired by the review mechanism of human-beings during the study. By leveraging such review mechanism, the performance of KD models can be substantially improved and in particular, the improved performance is generally better than that of the associated teacher networks, as shown by a series of experiments over six benchmark datasets. This demonstrates that the capacity gap issue could be effectively solved by the suggested review mechanism. We also show that the proposed review mechanism can be easily adapted to a kind of encoder-decoder based heavy crowd counting models to boost their performance without adding any additional model parameter. Moreover, we suggest a new variant of SKT model in the instruction phase of the proposed ReviewKD model, aiming to further improve the performance of the SKT model. A future direction is to establish some theoretical understandings on the success of the introduced review mechanism.

\section*{Acknowledgements}{
The work of J. Zeng  was supported in part by the National Natural Science Foundation of China [Grant No. 61977038] and by Thousand Talents Plan of Jiangxi Province [Grant No. jxsq2019201124]. 

 }

\bibliographystyle{IEEEtran}
\bibliography{reference}

\begin{thebibliography}{10}
\providecommand{\url}[1]{#1}
\csname url@samestyle\endcsname
\providecommand{\newblock}{\relax}
\providecommand{\bibinfo}[2]{#2}
\providecommand{\BIBentrySTDinterwordspacing}{\spaceskip=0pt\relax}
\providecommand{\BIBentryALTinterwordstretchfactor}{4}
\providecommand{\BIBentryALTinterwordspacing}{\spaceskip=\fontdimen2\font plus
\BIBentryALTinterwordstretchfactor\fontdimen3\font minus
  \fontdimen4\font\relax}
\providecommand{\BIBforeignlanguage}[2]{{%
\expandafter\ifx\csname l@#1\endcsname\relax
\typeout{** WARNING: IEEEtran.bst: No hyphenation pattern has been}%
\typeout{** loaded for the language `#1'. Using the pattern for}%
\typeout{** the default language instead.}%
\else
\language=\csname l@#1\endcsname
\fi
#2}}
\providecommand{\BIBdecl}{\relax}
\BIBdecl

\bibitem{ma2019bayesian}
Z.~Ma, X.~Wei, X.~Hong, and Y.~Gong, ``Bayesian loss for crowd count estimation
  with point supervision,'' in \emph{Proceedings of the IEEE International
  Conference on Computer Vision}, 2019, pp. 6142--6151.

\bibitem{jiang2020attention}
X.~Jiang, L.~Zhang, M.~Xu, T.~Zhang, P.~Lv, B.~Zhou, X.~Yang, and Y.~Pang,
  ``Attention scaling for crowd counting,'' in \emph{Proceedings of the IEEE
  Conference on Computer Vision and Pattern Recognition}, 2020, pp. 4706--4715.

\bibitem{thanasutives2021encoder}
P.~Thanasutives, K.-i. Fukui, M.~Numao, and B.~Kijsirikul, ``Encoder-decoder
  based convolutional neural networks with multi-scale-aware modules for crowd
  counting,'' in \emph{Proceedings of the International Conference on Pattern
  Recognition (ICPR)}.\hskip 1em plus 0.5em minus 0.4em\relax IEEE, 2021, pp.
  2382--2389.

\bibitem{topkaya2014counting}
I.~S. Topkaya, H.~Erdogan, and F.~Porikli, ``Counting people by clustering
  person detector outputs,'' in \emph{Proceedings of the 11th IEEE
  International Conference on Advanced Video and Signal Based Surveillance
  (AVSS)}.\hskip 1em plus 0.5em minus 0.4em\relax IEEE, 2014, pp. 313--318.

\bibitem{li2008estimating}
M.~Li, Z.~Zhang, K.~Huang, and T.~Tan, ``Estimating the number of people in
  crowded scenes by mid based foreground segmentation and head-shoulder
  detection,'' in \emph{Proceedings of the 19th International Conference on
  Pattern Recognition}.\hskip 1em plus 0.5em minus 0.4em\relax IEEE, 2008, pp.
  1--4.

\bibitem{leibe2005pedestrian}
B.~Leibe, E.~Seemann, and B.~Schiele, ``Pedestrian detection in crowded
  scenes,'' in \emph{Proceedings of the 2005 IEEE Computer Society Conference
  on Computer Vision and Pattern Recognition (CVPR'05)}, vol.~1.\hskip 1em plus
  0.5em minus 0.4em\relax IEEE, 2005, pp. 878--885.

\bibitem{enzweiler2008monocular}
M.~Enzweiler and D.~M. Gavrila, ``Monocular pedestrian detection: Survey and
  experiments,'' \emph{IEEE Transactions on Pattern Analysis and Machine
  Intelligence}, vol.~31, no.~12, pp. 2179--2195, 2008.

\bibitem{chan2008privacy}
A.~B. Chan, Z.-S.~J. Liang, and N.~Vasconcelos, ``Privacy preserving crowd
  monitoring: Counting people without people models or tracking,'' in
  \emph{Proceedings of the IEEE Conference on Computer Vision and Pattern
  Recognition}.\hskip 1em plus 0.5em minus 0.4em\relax IEEE, 2008, pp. 1--7.

\bibitem{idrees2013multi}
H.~Idrees, I.~Saleemi, C.~Seibert, and M.~Shah, ``Multi-source multi-scale
  counting in extremely dense crowd images,'' in \emph{Proceedings of the IEEE
  Conference on Computer Vision and Pattern Recognition}, 2013, pp. 2547--2554.

\bibitem{chan2009bayesian}
A.~B. Chan and N.~Vasconcelos, ``Bayesian poisson regression for crowd
  counting,'' in \emph{Proceedings of the 12th IEEE International Conference on
  Computer Vision}.\hskip 1em plus 0.5em minus 0.4em\relax IEEE, 2009, pp.
  545--551.

\bibitem{chen2012feature}
K.~Chen, C.~C. Loy, S.~Gong, and T.~Xiang, ``Feature mining for localised crowd
  counting.'' in \emph{Proceedings of the British Machine Vision Conference},
  vol.~1, no.~2, 2012, p.~3.

\bibitem{bai2020adaptive}
S.~Bai, Z.~He, Y.~Qiao, H.~Hu, W.~Wu, and J.~Yan, ``Adaptive dilated network
  with self-correction supervision for counting,'' in \emph{Proceedings of the
  IEEE Conference on Computer Vision and Pattern Recognition}, 2020, pp.
  4594--4603.

\bibitem{zhang2016single}
Y.~Zhang, D.~Zhou, S.~Chen, S.~Gao, and Y.~Ma, ``Single-image crowd counting
  via multi-column convolutional neural network,'' in \emph{Proceedings of the
  IEEE Conference on Computer Vision and Pattern Recognition}, 2016, pp.
  589--597.

\bibitem{li2018csrnet}
Y.~Li, X.~Zhang, and D.~Chen, ``Csrnet: Dilated convolutional neural networks
  for understanding the highly congested scenes,'' in \emph{Proceedings of the
  IEEE Conference on Computer Vision and Pattern Recognition}, 2018, pp.
  1091--1100.

\bibitem{miao2020shallow}
Y.~Miao, Z.~Lin, G.~Ding, and J.~Han, ``Shallow feature based dense attention
  network for crowd counting.'' in \emph{Proceedings of the AAAI Conference on
  Artificial Intelligence}, 2020, pp. 11\,765--11\,772.

\bibitem{sindagi2019ha}
V.~A. Sindagi and V.~M. Patel, ``Ha-cnn: Hierarchical attention-based crowd
  counting network,'' \emph{IEEE Transactions on Image Processing}, vol.~29,
  pp. 323--335, 2019.

\bibitem{walach2016learning}
E.~Walach and L.~Wolf, ``Learning to count with cnn boosting,'' in
  \emph{Proceedings of the European Conference on Computer Vision}.\hskip 1em
  plus 0.5em minus 0.4em\relax Springer, 2016, pp. 660--676.

\bibitem{wang2015deep}
C.~Wang, H.~Zhang, L.~Yang, S.~Liu, and X.~Cao, ``Deep people counting in
  extremely dense crowds,'' in \emph{Proceedings of the 23rd ACM International
  Conference on Multimedia}, 2015, pp. 1299--1302.

\bibitem{fu2015fast}
M.~Fu, P.~Xu, X.~Li, Q.~Liu, M.~Ye, and C.~Zhu, ``Fast crowd density estimation
  with convolutional neural networks,'' \emph{Engineering Applications of
  Artificial Intelligence}, vol.~43, pp. 81--88, 2015.

\bibitem{zhang2015cross}
C.~Zhang, H.~Li, X.~Wang, and X.~Yang, ``Cross-scene crowd counting via deep
  convolutional neural networks,'' in \emph{Proceedings of the IEEE Conference
  on Computer Vision and Pattern Recognition}, 2015, pp. 833--841.

\bibitem{yan2021crowd}
Z.~Yan, R.~Zhang, H.~Zhang, Q.~Zhang, and W.~Zuo, ``Crowd counting via
  perspective-guided fractional-dilation convolution,'' \emph{IEEE Transactions
  on Multimedia}, vol.~24, pp. 2633--2647, 2021.

\bibitem{dollar2011pedestrian}
P.~Dollar, C.~Wojek, B.~Schiele, and P.~Perona, ``Pedestrian detection: An
  evaluation of the state of the art,'' \emph{IEEE Transactions on Pattern
  Analysis and Machine Intelligence}, vol.~34, no.~4, pp. 743--761, 2011.

\bibitem{chen2020crowd}
J.~Chen, W.~Su, and Z.~Wang, ``Crowd counting with crowd attention
  convolutional neural network,'' \emph{Neurocomputing}, vol. 382, pp.
  210--220, 2020.

\bibitem{jiang2020density}
X.~Jiang, L.~Zhang, T.~Zhang, P.~Lv, B.~Zhou, Y.~Pang, M.~Xu, and C.~Xu,
  ``Density-aware multi-task learning for crowd counting,'' \emph{IEEE
  Transactions on Multimedia}, vol.~23, pp. 443--453, 2020.

\bibitem{wang2021interlayer}
M.~Wang, H.~Cai, J.~Zhou, and M.~Gong, ``Interlayer and intralayer scale
  aggregation for scale-invariant crowd counting,'' \emph{Neurocomputing}, vol.
  441, pp. 128--137, 2021.

\bibitem{simonyan2014very}
K.~Simonyan and A.~Zisserman, ``Very deep convolutional networks for
  large-scale image recognition,'' \emph{arXiv preprint arXiv:1409.1556}, 2014.

\bibitem{liu2020efficient}
L.~Liu, J.~Chen, H.~Wu, T.~Chen, G.~Li, and L.~Lin, ``Efficient crowd counting
  via structured knowledge transfer,'' in \emph{Proceedings of the ACM
  Multimedia}, 2020.

\bibitem{cao2018scale}
X.~Cao, Z.~Wang, Y.~Zhao, and F.~Su, ``Scale aggregation network for accurate
  and efficient crowd counting,'' in \emph{Proceedings of the European
  Conference on Computer Vision (ECCV)}, 2018, pp. 734--750.

\bibitem{shen2018crowd}
Z.~Shen, Y.~Xu, B.~Ni, M.~Wang, J.~Hu, and X.~Yang, ``Crowd counting via
  adversarial cross-scale consistency pursuit,'' in \emph{Proceedings of the
  IEEE Conference on Computer Vision and Pattern Recognition}, 2018, pp.
  5245--5254.

\bibitem{jiang2019crowd}
X.~Jiang, Z.~Xiao, B.~Zhang, X.~Zhen, X.~Cao, D.~Doermann, and L.~Shao, ``Crowd
  counting and density estimation by trellis encoder-decoder networks,'' in
  \emph{Proceedings of the IEEE Conference on Computer Vision and Pattern
  Recognition}, 2019, pp. 6133--6142.

\bibitem{wang2020mobilecount}
P.~Wang, C.~Gao, Y.~Wang, H.~Li, and Y.~Gao, ``Mobilecount: An efficient
  encoder-decoder framework for real-time crowd counting,''
  \emph{Neurocomputing}, vol. 407, pp. 292--299, 2020.

\bibitem{hinton2015distilling}
G.~Hinton, O.~Vinyals, and J.~Dean, ``Distilling the knowledge in a neural
  network,'' \emph{arXiv preprint arXiv:1503.02531}, 2015.

\bibitem{li2021reskd}
X.~Li, S.~Li, B.~Omar, F.~Wu, and X.~Li, ``Reskd: Residual-guided knowledge
  distillation,'' \emph{IEEE Transactions on Image Processing}, vol.~30, pp.
  4735--4746, 2021.

\bibitem{gao2020residual}
M.~Gao, Y.~Shen, Q.~Li, and C.~C. Loy, ``Residual knowledge distillation,''
  \emph{arXiv preprint arXiv:2002.09168}, 2020.

\bibitem{mirzadeh2020improved}
S.~I. Mirzadeh, M.~Farajtabar, A.~Li, N.~Levine, A.~Matsukawa, and
  H.~Ghasemzadeh, ``Improved knowledge distillation via teacher assistant,'' in
  \emph{Proceedings of the AAAI Conference on Artificial Intelligence},
  vol.~34, no.~04, 2020, pp. 5191--5198.

\bibitem{howard2018inverted}
A.~Howard, A.~Zhmoginov, L.-C. Chen, M.~Sandler, and M.~Zhu, ``Inverted
  residuals and linear bottlenecks: Mobile networks for classification,
  detection and segmentation,'' \emph{arXiv preprint arXiv:1801.04381}, 2018.

\bibitem{idrees2018composition}
H.~Idrees, M.~Tayyab, K.~Athrey, D.~Zhang, S.~Al-Maadeed, N.~Rajpoot, and
  M.~Shah, ``Composition loss for counting, density map estimation and
  localization in dense crowds,'' in \emph{Proceedings of the European
  Conference on Computer Vision (ECCV)}, 2018, pp. 532--546.

\bibitem{guerrero2015extremely}
R.~Guerrero-G{\'o}mez-Olmedo, B.~Torre-Jim{\'e}nez, R.~L{\'o}pez-Sastre,
  S.~Maldonado-Basc{\'o}n, and D.~Onoro-Rubio, ``Extremely overlapping vehicle
  counting,'' in \emph{Proceedings of the Iberian Conference on Pattern
  Recognition and Image Analysis}.\hskip 1em plus 0.5em minus 0.4em\relax
  Springer, 2015, pp. 423--431.

\bibitem{kingma2014adam}
D.~P. Kingma and J.~Ba, ``Adam: A method for stochastic optimization,'' in
  \emph{Proceedings of the International Conference for Learning
  Representations (ICLR'14)}, Banff, Canada, 2014, pp. 1--15.

\bibitem{liu2019counting}
L.~Liu, H.~Lu, H.~Xiong, K.~Xian, Z.~Cao, and C.~Shen, ``Counting objects by
  blockwise classification,'' \emph{IEEE Transactions on Circuits and Systems
  for Video Technology}, vol.~30, no.~10, pp. 3513--3527, 2020.

\bibitem{sam2020locate}
D.~B. Sam, S.~V. Peri, M.~N. Sundararaman, A.~Kamath, and V.~B. Radhakrishnan,
  ``Locate, size and count: Accurately resolving people in dense crowds via
  detection,'' \emph{IEEE Transactions on Pattern Analysis and Machine
  Intelligence}, vol.~43, no.~8, pp. 2739--2751, 2021.

\bibitem{wan2020kernel}
J.~Wan, Q.~Wang, and A.~B. Chan, ``Kernel-based density map generation for
  dense object counting,'' \emph{IEEE Transactions on Pattern Analysis and
  Machine Intelligence}, vol.~44, no.~3, pp. 1357--1370, 2022.

\end{thebibliography}

\end{document}